\documentclass{article}

\usepackage[final]{neurips_2022}

\usepackage[utf8]{inputenc} % allow utf-8 input
\usepackage[T1]{fontenc}    % use 8-bit T1 fonts
\usepackage{url}            % simple URL typesetting
\usepackage{booktabs}       % professional-quality tables
\usepackage{amsfonts}       % blackboard math symbols
\usepackage{nicefrac}       % compact symbols for 1/2, etc.
\usepackage{microtype}      % microtypography
\usepackage[table]{xcolor}  % loads also »colortbl«         % colors
\definecolor{plum}  {rgb}{.5,0,.5}
\definecolor{forest}  {rgb}{0,.4,0} 
\definecolor{midnight}  {rgb}{0,0,.5} 
\definecolor{gray}  {rgb}{.8,.8,.8} 
\definecolor{orange}{rgb}{.87,.414,.062}
\definecolor{dkorange}{HTML}{D24E23}
\definecolor{violet}{HTML}{674ea7}
\definecolor{brick}{rgb}{.7,0,0}
\definecolor{green}{HTML}{417e27}  
\definecolor{blue}{rgb}{0,0,.7}
\definecolor{maroon}{rgb}{0.52,0,0}
\definecolor{royalpurple}{rgb}{0.47,0.32,0.66}
\definecolor{tan}{rgb}{0.82,0.70,0.54}
\usepackage[pdftex,plainpages=false,pdfpagelabels,colorlinks=true,urlcolor=midnight,linkcolor=black!80,citecolor=forest]{hyperref}

\usepackage{xspace}

\newenvironment{squishlist}
{   \begin{list}{$\bullet$}
    { 
    \setlength{\itemsep}{0pt}
    \setlength{\parsep}{.5em}
    \setlength{\topsep}{0pt}
    \setlength{\partopsep}{0pt}
    \setlength{\leftmargin}{1.5em} 
    \setlength{\labelwidth}{1.5em}
    \setlength{\labelsep}{0.8em} } }
      {\end{list}}

\newcommand{\bX}{\mathbf{X}}
\newcommand{\bY}{\mathbf{Y}}

\newcommand{\bZ}{\mathbf{Z}}
\newcommand{\bV}{\mathbf{V}}
\newcommand{\bR}{\mathbf{R}}

\newcommand{\bC}{\mathbf{C}}
\newcommand{\bK}{\mathbf{K}}
\newcommand{\by}{\mathbf{y}}
\newcommand{\btilZ}{{\tilde{\bZ}}}

\def\bphi{{\boldsymbol{\phi}}}
\def\bSigma{\boldsymbol{\Sigma}}
\newcommand{\tilphi}{\tilde{\bphi}}
\newcommand{\hatphi}{\boldsymbol{\hat{\phi}}}
\newcommand{\bs}[1]{\boldsymbol{#1}}
\usepackage{mathtools}
\DeclarePairedDelimiter\norm{\lVert}{\rVert}%
\DeclarePairedDelimiter\ang{\langle}{\rangle}%

\DeclareMathOperator*{\argmin}{arg\,min}

\def\eande{{\sc Embed \& Emulate}\xspace}

\usepackage{amsmath,amssymb}
\makeatletter
\newsavebox\myboxA
\newsavebox\myboxB
\newlength\mylenA

\newcommand*\xoverline[2][0.75]{%
    \sbox{\myboxA}{$\m@th#2$}%
    \setbox\myboxB\null% Phantom box
    \ht\myboxB=\ht\myboxA%
    \dp\myboxB=\dp\myboxA%
    \wd\myboxB=#1\wd\myboxA% Scale phantom
    \sbox\myboxB{$\m@th\overline{\copy\myboxB}$}%  Overlined phantom
    \setlength\mylenA{\the\wd\myboxA}%   calc width diff
    \addtolength\mylenA{-\the\wd\myboxB}%
    \ifdim\wd\myboxB<\wd\myboxA%
       \rlap{\hskip 0.5\mylenA\usebox\myboxB}{\usebox\myboxA}%
    \else
        \hskip -0.5\mylenA\rlap{\usebox\myboxA}{\hskip 0.5\mylenA\usebox\myboxB}%
    \fi}
\makeatother

\usepackage{multirow}

\usepackage{subfig}
\usepackage{graphicx}
\usepackage[export]{adjustbox}
\usepackage{derivative}
\usepackage{amsthm}

\usepackage{comma}

\usepackage{titlesec}
\usepackage{algorithm, algorithmic}
\usepackage{amssymb}% http://ctan.org/pkg/amssymb
\usepackage{pifont}% http://ctan.org/pkg/pifont
\titlespacing*{\subsection}{0pt}{-0pt}{-2pt}
\titlespacing*{\section}{0pt}{-2pt}{-2pt}
\titlespacing*{\paragraph}{0pt}{0pt}{3pt}
\titleformat{\paragraph}[runin]
  {\normalfont\bfseries}{}{}{}[:]

\title{Embed and Emulate: Learning to estimate parameters of dynamical systems with uncertainty quantification}

\author{%
  Ruoxi Jiang \\
  Department of Computer Science\\
  University of Chicago\\
  Chicago, IL 60637 \\
  \texttt{roxie62@uchicago.edu} \\
  \And
  Rebecca Willett \\
  Department of Statistics and Computer Science\\
  University of Chicago\\
  Chicago, IL 60637 \\
  \texttt{willett@uchicago.edu} \\
}

\usepackage[capitalise]{cleveref}
\crefformat{section}{\S#2#1#3}
\crefformat{appendix}{\S#2#1#3}
\def\pfixed{p_{\bphi,{\rm fixed}}}
\def\pempb{p_{\bphi,{\rm empB}}}
\newcommand{\ave}[1]{\langle #1 \rangle}

\begin{document}

\maketitle

\begin{abstract}
This paper explores 
learning emulators for 
parameter estimation with uncertainty estimation of high-dimensional dynamical systems. We assume access to a 
computationally complex simulator that inputs a candidate parameter and outputs a corresponding multichannel time series. Our task is to accurately estimate a range of likely values of the underlying parameters. 
Standard iterative approaches necessitate
running the simulator many times, which is computationally prohibitive.
This paper describes a novel framework for learning feature embeddings of observed dynamics jointly with an emulator that can replace high-cost simulators for parameter estimation. Leveraging a contrastive learning approach, our method exploits intrinsic data properties within and across parameter and trajectory domains. On a coupled 396-dimensional multiscale Lorenz 96 system, our method significantly outperforms a typical parameter estimation
method based on predefined metrics and a classical numerical simulator, and with only 1.19\% of the baseline's computation time.
Ablation studies highlight
the potential of explicitly designing learned emulators for parameter estimation by leveraging contrastive learning.

% . The problem is to estimate underlying parameters given sequences of multiple observations and quantify estimates uncertainty. 
% Though traditional methods(ESM) using predefined objective functions have to balance between accuracy and computational cost and requires domain expertise, we propose a unified framework to obtain feature embeddings and learn an emulator replacing high-cost simulators. 
% with a fraction of running of simulators when comparing point estimates, quantified uncertainty and empirical computational cost. Comparing to our advanced supervised regression which lacks the ability to quantify uncertainty, our method shows huge advantage within a small training size.

\end{abstract}

\section{Introduction}

Physics-based simulators play a vital role in many domains of science and engineering, from energy infrastructure to atmospheric sciences. They  are frequently  critical for assessing risk and
exploring ``what if'' scenarios, which require running models many times \citep{snyder2019joint,beusch2020emulating,helgeson2021simpler,zhao2021global}.
However, the increasing complexity of operational computer models makes running such simulators a major challenge. 
Emulators (also known as surrogate models) are models trained to
mimic numerical simulations at a much lower computational cost, particularly for parameters or inputs that have not been simulated. 

This paper considers the problem of estimating the parameters of a physics simulation that provide the best fit to data. 
The estimation problem  can generally be thought of as a nonlinear inverse problem in which our goal is to estimate parameters $\bphi \in \mathbb{R}^{k}$ from a noisy multichannel time series $\bZ \in \mathbb{R}^{T \times d}$. 
$\bZ = H(\bphi; \bZ_0) + \bs\eta$,
where $H(\bphi; \bZ_0)$ represents running is a physics simulator with parameters $\bphi$ and initial condition $\bZ_0$ for $T$ time steps, and $\bs\eta \sim \mathcal{N}(0, \bs\Gamma)$ is observation noise. 
To ease the notation, we drop the initial condition $\bZ_0$ in $H(\bphi; \bZ_0)$.
We are particularly interested in complex simulators for which we do not have analytic expressions for $H$, evaluating $H(\bphi)$ is a computational bottleneck and we cannot readily compute its gradients. 

One important application of this problem arises in climate science, where climate scientists have spent decades developing sophisticated physics-based models corresponding to $H$, often implemented using large-scale software systems that solve complex systems of differential equations \citep{kay2015community}. In this case, evaluating $H$ can be computationally demanding. 
Furthermore, tools like physics-informed neural networks \citep{raissi2019physics} are inapplicable because we cannot compute losses that depend upon knowing the form of $H$.
Estimating the parameters of such models based on observational data is essential for accurate climate forecasting, and we typically seek not only point estimates of parameters, but also quantifiable uncertainty measures that allow us to forecast the full range of possible future outcomes. 
Parameter uncertainty quantification is vital here
\citep{cleary2021calibrate,souza2020uncertainty, hansen2022central}, especially when dealing with noisy observations 
% where estimation errors exists inherently in the systems
where a small error in $\hat \bphi$ might lead to a dramatically different forecasts of the dynamics; uncertainty quantification also plays a vital role in the decision-making process to prevent hazardous loss under tail events \citep{hansen2022central}. 

\everypar{\looseness=-1}

% \paragraph{\eande}
We explore an alternative framework in which we use our simulator to generate training samples of the form $(\bphi_i,\bZ_i)$ for  $i=1,\ldots,n$ and train a machine learning model for parameter estimation. 
Our approach, which we call \eande, jointly trains two neural networks. The first, $f_\theta$, maps an observation $\bZ$ to a low-dimensional embedding. The second, $\hat g_\theta$, emulates the map $g_\theta := f_\theta \circ H$, and so maps a parameter vector $\bphi$ to the same low-dimensional embedding space. The learned $\hat g_\theta$ may be used in place of $H$ within an optimization-based framework such as Ensemble Kalman Inversion (EnKI, \cref{sec:enki}).

More specifically, our  method uses the learned low-dimensional embedding to specify an objective function well-suited to parameter estimation without expert knowledge and the learned emulator as a surrogate for classical numerical methods used to compute $H$. Furthermore, inspired by empirical Bayes methods, our learned network suggests a natural mechanism for specifying the EnKI prior based on the trained network. 

\section{Related work}

\paragraph{Brief parameter estimation background}
\label{sec:enki}
A widely-used approach for parameter estimation is Ensemble Kalman Inversion (EnKI; \cref{section:appendix_enki}) \citep{Iglesias_2013}. 
EnKI is a derivative-free optimization framework
allows a user to specify a prior distribution over the parameter vector $\bphi$ and results in samples from the corresponding posterior distribution. As a result, this method provides users with important information about the uncertainty of estimates of $\bphi$. 
The clear and thought-provoking paper of \citet{schneider2017earth} 
highlighted the use of the EnKI in earth system modeling.

Despite its success in the perfect-model setting (i.e., where $H(\bphi)$ can be computed exactly and perfectly represents the dynamics in $\bZ$ for some $\bphi$),
using such methods in practice presents several challenges. To use the EnKI for parameter estimation, we must specify three key elements: (a) the objective function that EnKI is attempting to minimize; (b) the tool used to compute the forward mapping $H$ (or a surrogate for $H$); and (c) a prior $p_\bphi$. 
For instance, \citet{schneider2017earth} uses a Gaussian prior $p_\bphi$, computes $H$ using Runge-Kutta methods \citep{dormand1980family}, and attempts to minimize 
the Mahalanobis distance
\begin{equation}
    J_{\rm moment}(\bphi) :=  \| m(\bZ_i) - m( H(\bphi))\|_{\bSigma(m(\bZ_i))}^2, 
    \label{eq:moment}
\end{equation}
where $m(\bZ_i)$ is a vector of first and second moments of different spatial channels of $\bZ_i$ and $\bSigma_i$ is a diagonal matrix with the $j$-th diagonal entry $\bSigma_{i,j,j} := {\rm Var}[ m(H(\bphi_i))_j]$. Using a loss based on moments within the EnKI framework provides critical robustness to the chaotic nature of $\bZ_i$.

This framework presents two central challenges.
First, choosing the objective for EnKI to minimize 
requires non-trivial domain knowledge, and a poor choice may lead to biased parameter estimates or unpredictable sensitivities to certain features in $\bZ$.
Second, high-resolution simulations used to evaluate $H$ for a given $\bphi$ (e.g. Runge-Kutta or other classical numerical solvers) can be computationally demanding. In particular, the EnKI method uses a collection of {\em particles} to represent samples of the posterior, and the forward  model $H$ must be calculated for each particle at each iteration. For complex optimization landscapes or in high-dimensional settings, the number of particles  must be large. This is further complicated by the fact that often the parameter estimation task is conducted repeatedly, e.g., each time new climate data is acquired.

\textbf{Simulation-based inference (SBI):}
A large body of work in SBI 
% share the same interest with us regarding
focuses on parameter estimation for physics-based simulators \citep{beaumont2009adaptive,papamakarios2019sequential, cranmer2020frontier, lueckmann2019likelihood, chen2020neural, alsing2019fast}.
Advanced SBI methods focus on adaptive generation of new training data to approximate the posterior and could be classified into different categories based on how they adaptively choose informative simulations \citep{lueckmann2021benchmarking}. 
One approach, likelihood estimation,
proposes to approximate an intractable likelihood function \citep{drovandi2018approximating}.
For instance, Sequential Neural Likelihood estimation (SNL,  \citet{papamakarios2019sequential}) trains a conditional neural density estimator (e.g., Masked Autoregressive Flow (MAF)) which models the conditional distribution of data given parameters \citep{papamakarios2017masked}).
% to eventually approximate the likelihood.
While SNL does not readily scale to high-dimensional data, a recent variant called SNL+  \citep{chen2020neural} addresses this limitation by learning sufficient statistics (embeddings) of the data based on the infomax principle, and it iteratively updates the network used to compute sufficient statistics and the neural density estimators during sequential sampling.
An alternative approach, (sequential) neural posterior estimation ((S)NPE), aims to approximate directly the target posterior (SNPE-A in \citet{papamakarios2016fast}, SNPE-B in \citet{lueckmann2017flexible}, SNPE-C in \citet{greenberg2019automatic}), where SNPE-C forces fewer restrictions on the form of prior and posterior by leveraging neural conditional density estimation.

However, in the context of this manuscript's setting, i.e., estimating multiple $\boldsymbol{\phi}_i$ for multiple different observations $\mathbf{Z}_i$ at test time, 
the key technique of SBI, sequential sampling, can require substantial computational investments. This idea is discussed further with our experimental results. 
% becomes less appealing, which
% might harm their performance in terms of computational complexity with classical methods which do not depend on generating training data.

% Closely related to Papamakarios et al. [2019],  Lueckmann et al. [2019]  tries to ``emulate'' the simulator likelihood
% and advocates the use of active learning to choose the simulations in the next round based on uncertainty in the posterior. We are unable to compare with this method in this rebuttal as Lueckmann et al. [2019]
% does not appear to have available code and also lacks evidence that it can scale to high-dimensional data settings (their high-dimensional setting is of dimension $O(10^3)$ while ours is  $O(10^6)$). 

\paragraph{Learned emulators}
\label{sec:emulators}

Physics models and simulations are pervasive in weather and climate science, astrophysics, high-energy and accelerator physics, and the study of dynamical systems. These models and simulations are used, for example, to infer the underlying physical processes and equations that govern our observations, design new sensors or facilities, estimate errors, understand experimental behavior, and estimate unknown parameters within models. Many such physics simulators require expensive computational resources, and are difficult to fully leverage. 
``Learned emulators,'' ``surrogate models,'' or ``approximants'' are computationally-efficient approximate models that use numerically simulated data to train a machine learning system to mimic these numerical simulations at a much smaller computational cost. 
There have been many
recent successes in the development of surrogate model foundations \citep{brockherde2017bypassing,raissi2019physics,chattopadhyay2019data,chattopadhyay2020data,raissi2017physics,vlachas2018data,vlachas2020backpropagation,brajard2020combining,gagne2020machine} and applications \citep{goel2008surrogate,mengistu2008aerodynamic,brigham2007surrogate,kim2015time,papadopoulos2018neural,white2019multiscale,gentine2018could,rasp2018deep,cohen2020unified,yuval2020stable,xue2021predicting,wang2019fast}.

Typically, one would directly try to emulate $H$, and the efficacy of the learned emulator would be evaluated on how well $\hat H(\bphi)$ matches $H(\bphi)$ (often using squared error) across a range of $\bphi$. This formulation, however, may be suboptimal when the emulator will be used for parameter estimation.
First, as we will detail later, emulating $H$ when $\bZ$ is very high-dimensional is quite challenging. (In our experiments, ${\rm dim}(\bZ) = 396,000$.)
Furthermore, in many climate settings, $H$ represents a chaotic process in which very small changes to the initial conditions can result in very large differences later in the process. In this setting, training an emulator with a loss akin to $(1/n) \sum_{i=1}^n \|\bZ_i-\hat H(\bphi_i)\|_2^2$ may be overly sensitive to noise and initial conditions and not preserve statistical features of $\bZ_i$ that may be essential to parameter estimation.

Inspired by the objective in \eqref{eq:moment}, \citet{cleary2021calibrate} trained a model $\hat g_\theta$ to emulate the moments of parameters using the loss function $\ell_{\rm moment}(\theta) := (1/n) \sum_{i=1}^n 
\| m(\bZ_i) - \hat g_\theta(\bphi_i))\|_{\bSigma_i}^2$ to ease the burden of running $H$. However, $\ell_{\rm moment}(\theta)$ is not always the best choice.
First,  estimating $\bSigma_i$ for each training sample is an enormous computational burden that far exceeds the cost of generating the original training data (unless the range of candidate $\bphi$ is very small). Second,  choosing the moment function $m$ (essentially fixing a particular low-dimensional embedding) is a critical design element that requires domain expertise and often must be tuned in practice. Finally, even if one is willing to generate accurate estimates of $\bSigma_i$, one could face additional challenges. 
\citet{cleary2021calibrate} is designed for inferring parameters given one fixed test observation, and they train their emulator for a small domain of parameter space dependent on the test observation at hand. Extending this framework to a larger domain of parameters $\bphi$ results in large variability among the collection of corresponding $\bSigma_i$s generated for training, making the loss landscape challenging to navigate.

\paragraph{Contrastive learning}
With the above challenges in mind, we present a novel and generic framework for parameter estimation by jointly optimizing (a) an embedding of the multichannel dynamics used to evaluate the accuracy of a candidate parameter $\bphi$ and (b) 
an emulator of the dynamics projected into the embedding space.
We leverage a contrastive framework to learn discriminative representations for high-dimensional spatio-temporal data.

Contrastive representation learning has exploded in popularity recently for self-supervised visual feature learning that has achieved comparable performance with its supervised counterparts \citep{caron2020unsupervised, chen2020simple, he2020momentum, grill2020bootstrap, zhang2020self, caron2021emerging}. The majority of these frameworks operate
under the push-pull principle for instance-wise discrimination: images generated from different forms of data augmentation (e.g., cropping, color jittering) are pulled together while other images are pushed away. Apart from its common setup in unsupervised representation learning, \citet{khosla2020supervised} also demonstrated selecting positive and negative pairs in a supervised way leveraging label information can boost the performance. Many efforts focus on understanding the mechanism of contrastive learning \citep{oord2018representation, wei2020theoretical, arora2019theoretical, wang2020understanding, zimmermann2021contrastive}. Among them, \citet{zimmermann2021contrastive} show that contrastive loss can be interpreted as the cross-entropy between the latent conditional distribution and ground truth distribution.

\section{Contributions}

In our work, we follow the contrastive framework and learn an embedding network to capture structural information for high-dimensional spatio-temporal data and an emulator that can be used to compute parameter estimates with  uncertainty estimates.
This is effective for several reasons. 
First, when the model $H$ is bijective, 
representations learned using contrastive losses are highly correlated with their underlying parameters \citep{zimmermann2021contrastive}. 
Second, most existing emulating methods focus on approximating dynamical models for a fixed parameter $\bphi$ and lack generalization capacity \citep{krishnapriyan2021characterizing}. In part this is due to the difficulties of emulating dynamics with high spatio-temporal resolution. We circumvent this bottleneck by training an emulator to output
the learned latent representations of the dynamics, which are much lower-dimensional and easier to learn with fewer training samples. 
Finally, 
our method is inspired by the Contrastive Language–Image Pre-training (CLIP) \citep{radford2021learning} framework, which is designed for contrastive learning of aligned representations of images and language. Our approach, by analogy, learns aligned representations of dynamics and parameters, allowing us to jointly learn the embedding function and emulator for high-resolution data.

More specifically, this paper makes the following key contributions:
\begin{squishlist}
    \item Incorporation of a contrastive loss function for learning an embedding of the simulation outputs ($\bZ_i$) instead of relying on a known ``good'' moment function $m$, leveraging ideas from CLIP \citep{radford2021learning};
    % this approach  leveraging CLIP ideas
    \item Development of an emulator designed to facilitate parameter estimation with uncertainty quantification. We demonstrate that the proposed emulator can be used effectively within an EnKI framework to generate parameter estimates with accurate posterior distribution estimates. 
    \item An empirical evaluation of different methods (EnKI \citep{schneider2017earth}, supervised learning, NPE-C \cite{greenberg2019automatic}, SNL+ \citep{chen2020neural}, and our proposed emulator-based approach) in terms of 
    robustness to noise or errors in training and testing data for a range of sample sizes as well as
    computational costs.  
    Compared to using numerical solvers, we show that our method achieves higher accuracy overall in 1.08\% of the computation time, even accounting for the computational costs of generating training data. 
    \item Evaluation of the quality of the uncertainty estimates using the Continuous Ranked Probability Score (CRPS) for varying numbers of training samples.
\end{squishlist}

Our empirical results highlight both the improved computational and empirical accuracy of our proposed emulator-centric approach relative to competing methods and, more broadly, the benefits of designing emulators specifically for parameter estimation tasks.  

\section{Proposed method}
Our goal is to learn a mapping from observations of a multichannel dynamical system, $\bZ \in \mathbb{R}^{T \times d}$, to the underlying system parameters $\bphi \in \mathbb{R}^{k}$, where $\bZ = H(\bphi) + \bs\eta$ for some noise vector $\bs\eta \in \mathbb{R}^{T \times d}$. We assume we do not have access to an analytical expression for $H$, but can compute $H(\bphi)$ for any $\bphi$ using a computationally complex simulator. We further assume a prior $p_\bphi$ over parameter space.

Using this prior and the simulator, we  generate $n$ training samples as follows: for $i = 1,\ldots,n$, draw $\bphi_i \sim p_\bphi$, and use the simulator to compute $\bZ_i = H(\bphi_i) + \bs\eta_i$. 
We do not explicitly generate noise $\bs\eta_i$; rather, the numerical algorithm used to simulate $H$ will generally produce some errors, with faster implementations of $H$ being more error-prone. 
The distribution $p_\bphi$ coupled with the noise distribution over $\bs\eta$ induces the joint distribution $p_{\bphi,\bZ}$.
% \roxie{We assume the joint distribution $p(\bphi, \bZ)$.}

\subsection{Baseline approach -- supervised learning}
Given training samples $(\bphi_i,\bZ_i)$ for $i = 1,\ldots,n$, we train a neural network represented by $h_\theta$ using a ResNet \citep{he2016deep} as the backbone (see details in \cref{appendix:nn_architecture}). $\theta$ denotes the learned network parameters and our prediction is $\hat \bphi_i = h_\theta(\bZ_i)$. We train using the Mean Absolute Percentage Error (MAPE) loss
% $$
% \ell_{\rm supervised}(\theta) := 
% \frac{1}{n}\sum_{i=1}^n \| \bphi_i - h_\theta(\bZ_i) \|_2^2.
% $$
\begin{equation}
\ell_{\rm MAPE}(\theta) := 
\frac{1}{n}\sum_{i=1}^n \sum_{j=1}^k \left| \frac{\bphi_{ij} - (h_\theta(\bZ_i))_j }{|\bphi_{ij}| + \epsilon }\right|.
\label{eqn:loss_supervised}
\end{equation}
As we will see in the experimental results, this approach yields point estimates of $\bphi$ on holdout test data with reasonable accuracy and offers a significant computational advantage over the EnKI method. However, it does not offer uncertainty estimates, and its test accuracy 
% \roxie{I think yes. Its test accuracy (MAPE) is lower when training size is smaller. I am collecting CRPS results.}
is lower than that of our proposed approach (\cref{sec:ourApproach}).

\subsection{Our approach: joint embedding and emulation via contrastive  feature representation}
\label{sec:ourApproach}

% We study the parameter inference problem of a dynamical system, where we are given numerical data simulators (for example, complex climate models) $H(\cdot)$. 
% The data generator $H(\cdot)$ is assumed to be an injective function mapping from parameter space $\Phi \subset \mathbb{R}^{k}$ to trajectory space $\mathcal{Z} \subset \mathbb{R}^{d}$. 
% Given sequences of observations $\mathbf{Z} \in \mathbb{R}^{T \times d}$ over $T$ times, our goal to infer its parameters $\bphi$ by measuring how well $\bZ$ matches to $H(\hat{\bphi})$.

\textbf{Contrastive feature representation:}  Measuring the distance between a pair of dynamics, a necessary task for constructing a loss function used for training, is particularly challenging when  the dynamics are chaotic. 
% For example, simply using squared $L_2$ loss $\|\bZ -  H(\hat{\bphi})\|_2^2$ completely ignores the structural differences between
% a pair of $\bZ$s.
As mentioned above (\cref{sec:emulators}), one alternative  in the literature is to use a custom loss function based on summary  statistics of $\bZ$; this requires expert knowledge to determine which statistics are relevant, and  has demonstrated efficacy only for narrow ranges of parameters $\bphi$. 

We present an alternative framework in which we learn an embedding of $\bZ$, denoted $f_\theta(\bZ)$, 
 that  preserves statistical and structural  characteristics of $\bZ$ 
% so that the objective function
% \begin{equation}
% J(\bphi; \bZ, f_\theta, H) := \| f_\theta(\bZ) - f_\theta(H(\bphi) \|_2^2
% \label{eqn:J_obj}
% \end{equation}
% accurately reflects structural differences  between $\bZ_i$ and $\bZ_j$
\textit{most relevant to the parameter estimation task} and hence can be used to train an emulator. Here $\theta$ denotes all the parameters of the neural network defining the embedding function.  
% Identifying a characteristic function of $\bZ$ is then important. 
% Embracing the ``let-the-data-speak'' mentality promoted by \citet{hansen2022central} and 
Inspired by recent advances in contrastive representation learning, from a collection of parameters and trajectories pairs $\{\bphi_i, \mathbf{Z}_i\}_{i=1}^{n}$, we propose a trajectory encoder $f_\theta$ which learns a distinguishable representation by minimizing the following variant of the Info Noise Contrastive Estimation (InfoNCE) loss \citep{oord2018representation}: 
\begin{equation}
    \ell_{\bZ\bZ}(\theta; \tau) := \frac{1}{n} \sum\limits_{i=1}^{n} -\log 
    \frac{ \exp \big(%K(\bZ_i, \btilZ_i)
    \ang{f_\theta(\bZ_i),f_\theta(\btilZ_i)}
    /\tau\big) }
    { \sum\limits_{j=1}^{n} \exp ( %K(\bZ_i, \bZ_j) 
    \ang{f_\theta(\bZ_i),f_\theta(\bZ_j)} / \tau \big)}.
    \label{eqn:lcontra_f}
\end{equation}
Here $f_\theta(\bZ) \in \mathbb{S}^{p-1}$ is normalized and lives in a unit hypersphere. 
$\btilZ_i$ is selected through data augmentation (see \cref{appendix:implementation}) or in a supervised way by measuring the distance in the parameter space: $\btilZ_i := \argmin_{\bZ_j \neq \bZ_i} \delta(\bZ_j, \bZ_i)$, where $\delta$ is a metric used in calculating distance in parameter space, and
$\tau$ is a temperature hyperparameter balancing the impact of similar pairs and negative examples \citep{zhang2018heated, ravula2021inverse}. 
% \roxie{
In this way, we drive the model to embed data with similar values of the parameter $\bphi$ in similar locations while repulsing data with unrelated parameters, resulting in an embedding that respects the latent structure of the parameter and trajectory pairs. 
% \paragraph{Objective function } Given a learned feature representation $f_\theta$, our objective function measuring the distance between the observation $\bZ$ and simulation $H(\bphi)$ is defined as:

% \begin{equation}
%     J(\bphi; f, G, \bZ) = \| f_\theta(H(\bphi)) - f_\theta(\bZ) \|^2.
%     \label{eqn:J_obj}
% \end{equation}

\textbf{Emulators:} Emulating the dynamics of a simulator by training a deep neural network $\hat{g}_\theta(\cdot)$ is a natural approach to easing the computational burden of parameter estimation.
% burden of computational costs in running simulations $H(\cdot)$.
However, recent works in emulating dynamics focus on the fixed-parameter setting \citep{krishnapriyan2021characterizing} and cannot be easily generalized to multi-parameter settings (i.e. the emulator only approximates $H(\bphi)$ for a single fixed $\bphi$ instead of a range of $\bphi$s, as we need for parameter estimation). 

Learning emulators for a range of parameter values
is challenging in part due to the dimension of the dynamics to be emulated (i.e. the dimension of the $\bZ$s). 
% Training ``recurrent'' emulators 
% (i.e., instead of emulating the mapping $\bphi \rightarrow \{\bZ_t\}_{t=1}^T$, emulating the mapping $(\bphi, \bZ_{t-1}) \rightarrow \bZ_t$ and apply it $T$ times)  does not mitigate the challenge of high-dimensionality because optimizing the parameters of recurrent emulators is quite challenging. 
% \roxie{modified (highlighted for easier notice): And the problems become harder when we try to estimate $\bZ_t$ in a recurrent way up to $T$ times to capture the intrinsic temporal dynamics.} \todo{i don't understsand what you mean} \roxie{ you wrote before that we need to add another argument that emulating truncation of trajectories could not help ease the problem. So I thought it is not helpful in that (1)small truncation is useless; (2)large truncation in a recurrent way is very difficult to optimizer.}
To address this problem, we propose to learn an emulator $\hat{g}_\theta$ of $g_\theta := f_\theta \circ H$, 
which represents the mapping from the parameter space ($\bphi$) to the latent representation space $f_\theta(\bZ)$.
% \todo{We should NOT use $\hat G$ here because it suggests the output is the same dimension as $\bZ$. Let's denote the emulator by $\hat H_\theta$, which is emulating the mapping $H_\theta := f_\theta \circ G$. Note that $\theta$ denotes all learned parameters of our framework, including the weights of the embedding network $\hat f_\theta$ as well as the weights of the emulator network $\hat H_\theta$}
which is much lower-dimensional than $\bZ$ and hence easier to learn with limited training samples.
% \roxie{do I need to use $\hat G_{\theta'}$ to make it symmetric to $f_\theta$, is $\theta'$ a good notation?}

We say that the trajectory encoder and the emulator are \textit{perfectly aligned} when  $\hat{g}_\theta(\bphi)  = f_\theta(\bZ), \forall ~(\bphi, \bZ) \sim p_{\bphi, \bZ}$. Under perfect alignment, given a test $\bZ$, we may
seek to minimize
% our hope is that minimizing $J(\bZ_i; \bZ_j, f, G)$ defined in \eqref{eqn:J_obj} is equivalent to minimizing
% \todo{not true. If both $\hat H$ and $\hat f$ map to 0, you'd have perfect alignment but it would not be equivalent}
\begin{equation}
    J(\bphi; \bZ,f_\theta, \hat{g}_\theta) = \| \hat g_\theta (\bphi) - f_\theta(\bZ) \|^2_2 
    \label{eqn:J_obj_emu}
\end{equation}
% \todo{the use of $G$ vs $\hat G$ and $f_\theta$ vs $\hat f$ is sloppy. mistkaes like this make reviewers think this is thrown together thoughlessly and will get it rejected.}
instead of \eqref{eq:moment} with much faster objective function evaluations. 
% only \eqref{eqn:J_obj_emu} can be computed much more rapidly than \eqref{eqn:J_obj}. For instance, 

\textbf{Learning the emulator via CLIP:} 
Pretraining an embedding $f_\theta$ and then learning an emulator $\hat g_\theta$ with an $L_2$ loss 
is hard to optimize; the different architectures, necessitated by the very different dimensions of $\bphi$ and $\bZ$, make feature alignment  challenging and the optimization prone to poor local minimizers. 
% There are two possible methods of learning the emulator $\hat g_\theta (\cdot)$. First, if we had a trained feature embedding, we could potentially use $L_2$ loss and minimize $\ell_{\rm mse}(\theta) = (1/n)\sum_{i=1}^n \|\hat{g}_\theta(\bphi_i) - f_\theta(\bZ_i)\|^2$. However, we do not have a 
% clear mechanism for pre-training a 
% fixed feature embedding $f_\theta$, and jointly optimizing $\hat g_\theta$ and $f_\theta$ with respect to the $L_2$ loss presents significant numerical challenges. 
% our empirical experiments show such attempts encounter difficulties in convergence during the training procedure.
Alternatively, inspired by the recent success of CLIP \citep{radford2021learning, ravula2021inverse} in cross-domain representation learning, we use the following variant of InfoNCE loss to align the representation space between two networks:
\begin{equation}
    \ell_{\bZ\bphi}(\theta; \tau') := \frac{1}{n} \sum\limits_{i=1}^n -\log 
    \frac{ \exp \big(
    % K_{\bZ\bphi}(\bZ_i, \bphi_i) /\tau'
    \ang{f_\theta(\bZ_i),\hat{g}_\theta(\bphi_i)} / \tau'
    \big) }
    { \sum\limits_{j=1}^{n} \exp \big(
    % K_{\bZ\bphi}(\bZ_i, \bphi_j)
    \ang{f_\theta(\bZ_i),\hat{g}_\theta(\bphi_j)}
    / \tau' \big)} 
    - \log 
    \frac{ \exp \big(
    % K_{\bZ\bphi}(\bZ_i, \bphi_i)/\tau'
    \ang{f_\theta(\bZ_i),\hat{g}_\theta(\bphi_i)} / \tau'
    \big) }
    { \sum\limits_{j=1}^{n} \exp \big(
    % K_{\bZ\bphi}(\bZ_j, \bphi_i)
    \ang{f_\theta(\bZ_j),\hat{g}_\theta(\bphi_i)}
    / \tau' \big)} .
    \label{eqn:lcontra_gf}
\end{equation}
% \todo{you're using $f_\theta$ here, but I thought we were learning $\hat G$. } \roxie{Ultimately we are learning f and $\hat G$ together. But yes, the story I tell here is focusing on learning $\hat G$. I talked about unified training framework at the last paragraph. And if we are learning $\hat G$, the standard CLIP way is use only the left term in the above equation, i.e., fix traj and vary parameter. Should I add my explanation here?} 

\begin{figure}[t]
    \centering
    \subfloat[Network setup]{\includegraphics[width=.4\linewidth]{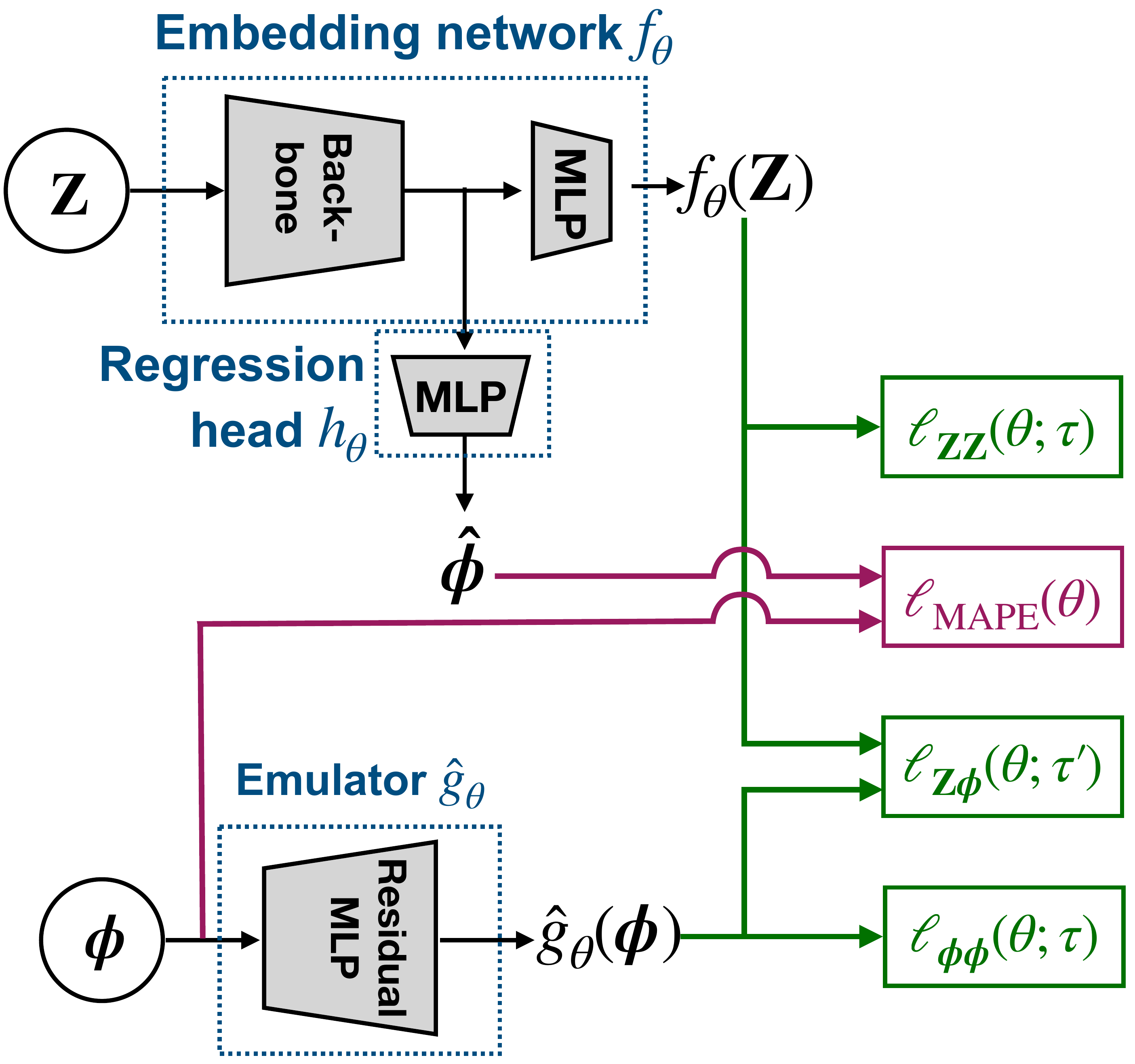}}\hfill
    \subfloat[Contrastive losses]{
    \includegraphics[width=.5\linewidth]{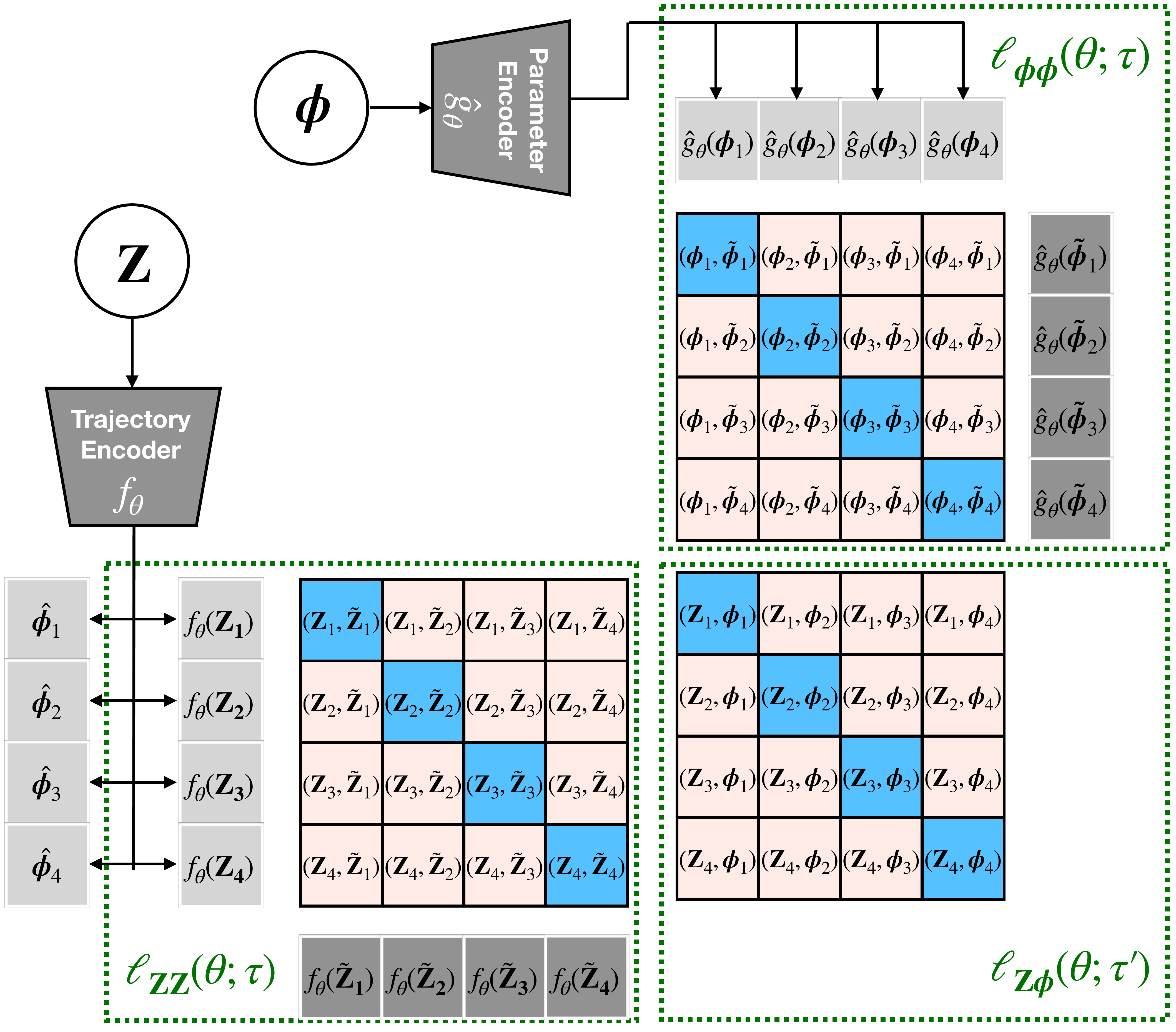}}
    \caption{\textbf{The \eande framework.} (a) Components of our network and loss function. 
    (b) We use three contrastive learning schemes. The bottom left block denotes the intra-trajecotry domain contrastive mechanism, the upper right denotes the intra-parameter domain contrastive learning, and the bottom right block shows inter-trajectory-parameter domains contrastive learning. Within each block, diagonals correspond to the dot product between the representations of positive trajectories pairs $(\bZ_i, \btilZ_i)$, positive parameter pairs $(\bphi_i, \tilphi_i)$, and matched trajectory parameter pairs $(\bZ_i, \bphi_i)$, which we aim to maximize. Off-diagonal pale pink blocks show similarities between the anchor point and the negatives examples, which we aim to minimize.}
    \label{fig:diagram_framework}
    \vspace{-.1in}
\end{figure}

\textbf{Intra-parameter domain contrastive loss:} In order to fully exploit data potential within parameter space and better guide the learning of the representation and preserve similarities within the parameter space, we add an intra-parameter contrastive loss and find empirically that it helps  accelerate the training process: 
% To better guide the learning of the emulator and enhance the discriminative ability of the representation, we add a parameter contrastive loss in a similar way to Eqn.\ref{eqn:lcontra_f}:
\begin{equation}
    \ell_{\bphi\bphi}(\theta; \tau) := \frac{1}{n} \sum\limits_{i=1}^n -\log 
    \frac{ \exp \big(
    % K_{\bphi, \bphi}(\bphi_i, \tilphi_i)
    \ang{\hat{g}_\theta(\bphi_i),\hat{g}_\theta(\tilphi_i)}
    /\tau \big) }
    { \sum\limits_{j=1}^{n} \exp \big(
    % K_{\bphi, \bphi }(\bphi_i, \bphi_j)
     \ang{\hat{g}_\theta(\bphi_i),\hat{g}_\theta(\bphi_j)}
    / \tau \big)},
    \label{eqn:lcontra_g}
\end{equation}
where $\tilphi$ is selected through data augmentation by perturbing $\bphi$ with small amount of noise (see  \cref{appendix:implementation}) or 
by finding nearest neighbors in the training set.
% in a supervised way: $\tilphi_i := \arg\min_{\bphi:= \bphi_j}\delta(\bphi_i, \bphi_j)$.

\textbf{Regression head:} \citet{zimmermann2021contrastive} shows that under certain assumptions (see  \cref{appendix:theoretical_analysis}), the feature encoder $f_\theta$ implicitly learns to invert the data generating process $H$ up to an affine transformation, implying we can estimate $\bphi$ via $h(f_\theta(\bZ))$ for some linear function $h$.
% \todo{do we want $H$ here or $g_\theta := f_\theta \circ H$?} \roxie{not exactly, we want $\hat{g}_\theta = f_\theta = A \circ H^{-1},$ A is an affine transformation}   
Inspired by their analysis, we add
a regression head $h_\theta$ (a single fully-connected linear layer) in our framework, and we ensure the affine relationship between $h_\theta$ and $f_\theta$ by letting them share the same backbone up to the last layers before the output. 
% We define the loss of $h_\theta$ as:\roxie{Or just say: we let $\ell_{\rm{regress}}(\theta)$ to be equal to the $\ell_{\rm supervised}(\theta)$ defined in \ref{eqn:loss_supervised}}.
% $$\ell_{\rm{regress}}(\theta) := \frac{1}{n}\sum_{i=1}^n \sum_{j=1}^k \left| \frac{\bphi_{ij} - (h_\theta(\bZ_i))_j }{|\bphi_{ij}| + \epsilon }\right| $$

\textbf{Unified training procedure:}  To this end, we formulate the final loss of our \eande method as
$$\ell(\theta) = \lambda_{\bZ\bZ} \ell_{\bZ\bZ}(\theta; \tau) + \lambda_{\bphi\bphi} \ell_{\bphi\bphi}(\theta; \tau) + \lambda_{\bZ\bphi}\ell_{\bZ\bphi}(\theta; \tau') + \lambda_{\rm MAPE}\ell_{\rm MAPE}(\theta),$$
where 
% $\ell_{\rm MAPE}$ is defined in \eqref{eqn:loss_supervised},
$\lambda_{\bZ\bZ}, \lambda_{\bphi\bphi}, \lambda_{\bZ\bphi}$ and $ \lambda_{\rm MAPE} $ control relative loss weights.
We show our basic setup in
\cref{fig:diagram_framework}(a)
 and loss framework in \cref{fig:diagram_framework}(b).

\section{Experiments}
\label{sec:experiment}
We conduct a numerical case study on the multiscale Lorenz-96 (L96) model \citep{lorenz1996predictability}, which is
a common test model for climate models with both ``fast'' (high-frequency) and ``slow'' (low-frequency) components and other geophysical applications \citep{majda2012filtering,law2012evaluating,law2016filter,brajard2020combining}. It is a prototypical turbulent dynamical system
 ``designed to mimic baroclinic turbulence in the midlatitude atmosphere'' \citep{majda2010mathematical}. 
Its dynamics exhibit strong energy-conserving non-linearities, and for some settings of $\bphi$, it can exhibit strong chaotic turbulence. {\def\UrlFont{\sf}
\def\UrlFont{\rm\small\ttfamily}
Code is available at: \url{https://github.com/roxie62/Embed-and-Emulate} 
}
The governing equations of the L96 system are defined as follows:
\begin{eqnarray*}
    \odv{\mathbf{X}^k_t}{t} = -\mathbf{X}^{k-1}_t(\mathbf{X}^{k-2}_t - \mathbf{X}^{k+1}_t) - \mathbf{X}^k_t + F - hc \bar{\mathbf{Y}}^k_t, \\
    \frac{1}{c} \odv{\mathbf{Y}^{j,k}_t}{t} = -b\mathbf{Y}^{j+1,k}_t(\mathbf{Y}^{j+2,k}_t - \mathbf{Y}^{j-1,k}_t) - \mathbf{Y}^{j,k}_t + \frac{h}{J}\mathbf{X}^k_t,
\end{eqnarray*}
where $\mathbf{X}_t \in \mathbb{R}^{K}$ denotes the slow variable at the $t$-th time stamp and $\mathbf{Y}_t \in \mathbb{R}^{KJ}$ denotes the fast variable. We use $\mathbf{Z}_t: = [\mathbf{X}_t, \mathbf{Y}_t] \in \mathbb{R}^{K(J+1)}$ to denote the system state at the $t$-th time stamp. We choose $K=36$ and $J=10$ throughout experiments in this section, as in \citet{schneider2017earth}.

% \paragraph{Baseline method }
We set our baseline following \citet{schneider2017earth}. Within the EnKI framework, they use Runge-Kutta \cite{dormand1980family} to compute the forward model $H$ and optimize \eqref{eq:moment}.
After analyzing the physics information of the L96 system, they define the moment function as:
$
m(\bZ) := [\ave{\bX}_T, \ave{\bar{\bY}}_T, \ave{\bX^2}_T, \ave{\bX\bar{\bY}}_T, \ave{\bar{\bY}^2}_T]
$, where $\ave{\cdot}_T$ denotes an empirical average over $T$ time steps, $\bar{\bY}$ denotes an average across the $J$ $\bY$ channels, 
$\ave{\bar{\bY}}_T = (1/JT) \sum_{t=1}^T \sum_{j=1}^J \bY_t^{j,k} \in \mathbb{R}^{K}$, and other quantities are computed similarly (see details in \cref{appendix:implementation}). 
$m(\bZ_t) \in \mathbb{R}^{5K}$. For the SBI methods, to suit the scenario we care about, i.e., estimating multiple $\boldsymbol{\phi}_i$ for multiple different observations $\mathbf{Z}_i$ at test time, we compare with the non-adaptive counterpart of the algorithms referenced in these papers. For instance, we set the round number of SNL+ \citep{chen2020neural} to be one. And use the fixed training dataset for SNL+ and NPE-C \citep{greenberg2019automatic}.

We conduct experiments and demonstrate the efficacy when measuring (1) the accuracy of averaged point estimates, (2) the accuracy of uncertainty quantification, (3) the empirical computational complexity.
For all of our experiments (including SNL+ and NPE-C), 
we use ResNet-34 \citep{he2016deep} as the backbone of the trajectory encoder. 
We apply average pooling at the last layer to generate a 512-d hidden vector. For contrastive learning, we project the 512-d hidden vector into a 128-d feature vector as the output of the trajectory encoder. For the regression head, we map the 512-d hidden vector to a 4-d parameter vector. It is important to note that activation functions are not used in the regression head to ensure an affine mapping between two projections. We parameterize the emulator network with a ResNet backbone, and to enhance the representation power of the network, we use five residual connection blocks (see \cref{appendix:nn_architecture}). We find this increases the alignment between the two representation spaces. 
Within each skip block, there is a residual connection between the input layer and the output. The implementation details for training are shown in  \cref{appendix:implementation}.

%-------------------------------------------
\subsection{Higher quality estimates with lower sample cost}
\label{section:vary_train_size}
In this section, we evaluate our method with different training sizes in the perfect-model setting where no noise is injected in observations. These experiments demonstrate the usefulness of our method in  realistic scenarios where forward models are prohibitively expensive and only limited quantities of training data are available. 

% \paragraph{Setup} 
We train our \eande method and the supervised regression baselines when the training size equals  $500$ and $1,000$. The training samples are sampled uniformly from $\bphi_{\min} = [F_{\min}, h_{\min}, c_{\min}, b_{\min}] = [-5, 0, 0.1, 0]$ to $\bphi_{\max} = [F_{\max}, h_{\max}, c_{\max}, b_{\max}] = [20, 5, 25, 25]$, and each simulation is of length 100, with $dt = 0.1$. We then sample 200 testing samples uniformly from a narrower range of $[-3, 0.5, 2, 2]$ to $[18, 4.5, 23, 23]$. As assumed in \citet{schneider2017earth}, each testing sample is of length $1,000$, with $dt = 0.1$. 
The EnKI prior used for baselines (i.e. minimizing \eqref{eq:moment}) is a Gaussian distribution with means at the middle of the range of the testing instances, diagonal covariance matrix, and variances broad enough that all test samples are within $2\sigma$ of the mean, denoted $\pfixed$ (see \cref{appendix:implementation}). 
Within the context of the \eande approach, we adopt an empirical Bayes approach: for each test instance, we compute $\hat \bphi$ using the regression head and use these values as the prior means when using EnKI to minimize \eqref{eqn:J_obj_emu}; we denote this prior  $\pempb$. 
The impact of the empirical Bayes approach is explored in \cref{sec:ablation}.

\begin{table}[ht]
  \centering
    %   \rowcolors{2}{gray!60}{white}
  \begin{tabular}{p{0.9em}p{10.85em}p{5.29em}p{5.29em}p{5.29em}p{5.29em}}
   %{clccccc}
    \toprule
     { $n$} & & {$F$  $\downarrow$} & {$h$  $\downarrow$} & {$c$  $\downarrow$} & {$b$  $\downarrow$} 
    \\
    \midrule
{\small 0} & {\small EnKI \textit{w/o} Learning}  & 15.48 (3.77) &  0.86 (0.20) & 40.45 (13.39) &  4.60 (0.60) \\
\midrule
 \multirow{5}{*}{{\small 500}} 
& {\small $h_{\theta}$ \textit{w/} \eande}   &  11.19  (3.65) & \textbf{3.18} (\textbf{1.60}) & \textbf{15.52} (6.24)  & \textbf{8.57 (2.17)} \\
& {\small EnKI \textit{w/} \eande} & \textbf{10.94} (4.07) & 3.74 (2.26) & 16.09 (6.41) & 8.84 (2.93) \\
& {\small Supervised Regression}   & 11.97 (\textbf{3.10}) & 4.07 (2.24) & 17.04 (\textbf{5.88})  & 9.07 (2.74) \\
& {{\small NPE-C}}  & 15.51 (4.52) &  5.94 (2.59) & 23.54 (8.16) & 9.42 (3.70) \\
& {\small SNL+}  & 36.88(19.93) &  48.71 (30.69) & 57.45 (28.59) & 23.45 (17.62) \\
  
\midrule
\multirow{5}{*}{{\small 1000}} 
&  {\small $h_{\theta}$ \textit{w/} \eande}   &  \textbf{6.30} (1.86) & \textbf{2.07} \textbf{(1.31)} & \textbf{9.34} (3.71) & \textbf{5.51} \textbf{(1.74)} \\
&  {\small EnKI \textit{w/} \eande} & 6.59 (2.14) & 2.36 (1.54) & 9.38 (4.02) & 5.54 (2.35) \\
&  {\small Supervised Regression} & 7.57 (\textbf{1.78})  & 3.08 (1.54)  & 11.46 (\textbf{3.29})  & 6.64 (2.19) \\
&  {{\small NPE-C}}  & 11.29 (3.54) & 5.62 (2.29) & 16.32 (6.53) & 6.81 (2.38) \\
&  {\small SNL+}  & 33.29(19.68) &  49.05 (27.75) & 48.17 (23.90) & 25.40 (16.03) \\
\bottomrule
\end{tabular}
 \vspace{.1in}
\caption{\textbf{Averaged MAPE (MdAPE, median absolute percentage error) for varying training size of different methods for 200 test samples.} 
We compare \eande $\textit{w/}$ regression head ($h_\theta$) and \eande plugged into EnKI to supervised regression, NPE-C \citep{greenberg2019automatic}, SNL+ \citep{chen2020neural}, and
a classical numerical solver (Runge-Kutta) plugged into EnKI to minimize \eqref{eq:moment}. 
The example illustrates that \eande is able to achieve a lower error
than both the EnKI approach based on a fixed moment vector objective and classical numerical solver \citep{schneider2017earth} and  a straightforward supervised regression approach that is unable to produce uncertainty estimates.}
\label{Table:mape}
% \vspace{-.3in}
\end{table}

\begin{table}[ht]
  \centering
    %   \rowcolors{2}{gray!60}{white}
  \begin{tabular}{p{12em}||p{5em}p{4em}p{4em}p{7em}}
   %{clccccc}
    \toprule
     & {\small Training data generation} & {\small Training time} & {\small 200 test runs} & {\small Total time (train + test)} 
    \\
    \midrule
{\small EnKI \textit{w/o} Learning}  & 0.0 & 0.0  & 8,000.0  & 8,000 (5.5 d) \\
{\small $h_\theta$  \textit{w/} \eande}   &  21.0 & 72.0 & 1.0 & 94.0 (1.57  h) \\
{\small EnKI  \textit{w/} \eande}   &  21.0 & 72.0 & 2.0 & 95.0 (1.58  h) \\
{\small Supervised Regression}   & 21.0 &  59.0 & 1.0 & 81.0 (1.35 h) \\ % \todo {old {\small NPE-C}} & 20.0 & 130.0 & 2.0 & 152.0 (2.53 h) \\
{\small NPE-C} & 21.0 & 72.0 & 2.0 & 95.0 (1.58 h) \\
SNL+   & 21.0 &  73.0 & 400.0 & 494.0 (8.23 h) \\
\bottomrule
\end{tabular}
\vspace{.1in}
\caption{\textbf{Empirical computational time for different stages of different methods} ($n=1,000$). Reported in minutes, total time for \eande, the supervised regression, or NPE-C are $1.19\%$ of EnKI with Runge-Kutta. All neural approaches are trained with 4 GPUs and tested with 1 GPU. Both training data generation and EnKI \textit{w/o} Learning are run with 32 CPU Cores.
  }
\label{table:computational_time_l96}
\vspace{-.2in}
\end{table}

We first evaluate the accuracy of point estimates using the mean and median absolute percentage error (MAPE and MdAPE), see \cref{appendix:implementation}.
In \cref{Table:mape} and \cref{table:computational_time_l96}, we see that
\eande evaluated with the regression head ($h_\theta$) and EnKI guarantees similar performance in terms of averaged accuracy. However, the regression head ($h_\theta$) is slightly better, which may be explained by our empirical observation that when estimates from 
$h_\theta$ with \eande are far from the true parameter values, then using this estimate as the prior mean for EnKI can worsen the estimate. Despite this challenge, the EnKI has the advantage of providing uncertainty estimates, which are discussed below and reflected in \cref{table:CRPS}.

Compared with other methods, it is clear that 
\eande yields a significant improvement in accuracy, especially for the parameter affecting high-frequency dynamics (e.g., $c$).
In particular, NPE-C 
\citep{greenberg2019automatic} 
performs worse with a smaller training set size % and requires a longer training time to converge
;
SNL+ 
\citep{chen2020neural}
using  MCMC or rejection sampling suffers from increasing evaluation time as the number of test samples increases. 
Moreover, when compared to the classical method of running
EnKI using a predefined moment function and expensive numerical solvers, \eande yields better performance in terms of accuracy and computation time. 
Last, our \eande framework also performs well relative to supervised regression for smaller training sets. 

We then evaluate our results based on the continuous ranked probability score (CRPS) \citep{hersbach2000decomposition, zamo2018estimation, pappenberger2015know}. CRPS is an important metric used in quantifying uncertainty (e.g., in weather forecasts). It measures the accuracy of estimated posterior distributions and
is defined as
$$
    {\rm CRPS}(C, \bphi_{*j}) = \int^{\infty}_{-\infty}(C(y_j) - U(y_j - \bphi_{*j}))^2 {\rm d}y_j,
$$
where $\hatphi_{*j}$ represents the $j$-th component of the estimate parameter vector, $\bphi_{*j}$ represents the $j$-th component of the true parameter vector, $C$ is the cumulative density function of the ensemble estimates with $C(y) = P(\bphi_{*j} \leq y)$, and $U$ is the  Heaviside step function.
For a deterministic estimate from supervised regression, CRPS is equal to the mean absolute error (MAE). \cref{table:CRPS} shows that \eande achieves high-accuracy estimates of uncertainty. 

\begin{table}[ht]
  \centering
   \begin{tabular}{p{1.1em}p{12em}p{4.5em}p{4.5em}p{4.5em}p{4.5em}}
\toprule
$n$ & & {$F$  $\downarrow$} & {$h$  $\downarrow$} & {$c$  $\downarrow$} & {$b$  $\downarrow$} \\
\midrule
{\small 0} & {\small EnKI \textit{w/o} Learning}  & 0.910 & 0.019 & 2.443 & 0.393 \\
\midrule
\multirow{4}{*}{\small 500} 
& $h_{\theta}$ \textit{w/} \eande   &  0.698 & 0.076 & 1.715 & 0.824 \\
& {\small EnKI \textit{w/} \eande} & \textbf{0.615} & \textbf{0.073} & \textbf{1.561} & \textbf{0.720}  \\
& {\small Supervised Regression}  & 0.707 & 0.104 & 1.785 &  0.917   \\
& {{\small NPE-C}}  & 0.844 & 0.106 & 2.117 & 0.853 \\
& {\small SNL+}  & 1.399 &  0.616 &  2.426 &  1.822 \\
  
\midrule
\multirow{4}{*}{\small 1000} 
& $h_{\theta}$ \textit{w/} \eande   &  0.412 & 0.049 &  0.992 & 0.453   \\
& {\small EnKI \textit{w/} \eande} & \textbf{0.360} & \textbf{0.042} & \textbf{0.829} &  \textbf{0.394}  \\
& {\small Supervised Regression}   & 0.478 &  0.078 &  1.242 & 0.650  \\
& {{\small NPE-C}}  & 0.593 & 0.096 & 1.389 & 0.564 \\
& {\small SNL+}  & 1.304 &  0.595 &  2.010 &  1.763 \\
\bottomrule
\end{tabular}
\vspace{.1in}
\caption{\textbf{Continuous Ranked Probability Score (CRPS) evaluated on 200 test samples.} The errors of the uncertainty estimates are almost always lower for \eande than for an EnKI method using a classical numerical solver.
And compared to neural methods, EnKI with \eande yields significant improvements over a supervised regression which cannot quantify uncertainty, defeats the simulation-based inference models NPE-C \citep{greenberg2019automatic} and SNL+ \citep{chen2020neural} which relies on sequential sampling.
Moreover, it's clear that \eande evaluated with EnKI provides more accurate uncertainty estimates than  point-wise estimates from the regression head ($h_\theta)$ trained with \eande. }  
\label{table:CRPS}
\vspace{-.1in}
\end{table}

%-------------------------------------------
\subsection{Visualizing uncertainty with noisy observations}
\label{sec:exp_noisy}
In this subsection, we go beyond the perfect-model setting and evaluate our method in the realistic scenarios where observations are noisy and obtained through: 
$\bZ = H(\bphi) + \bs\eta$, where $\bs\eta \sim \mathcal{N}(0, r\bs\Gamma)$, where $\bs\Gamma$ is the temporal covariance of the trajectory $\bZ$, and $r$ is a scaling value. Specifically, we set $\bphi = [10, 1, 10, 10]$ following \citet{schneider2017earth} to ensure we are in the chaotic regime. We use model trained with $n=4,000$ data samples and compare the results obtained in both the noiseless and noisy cases. 
For this visualization, both methods use the fixed prior from \cref{section:vary_train_size} based on the range of test values, so any differences observed are not caused by differences in the prior, but rather differences in the choice of objective and the method of computing the forward model. 
\begin{figure}[ht]
    \centering
    \includegraphics[width = \textwidth]{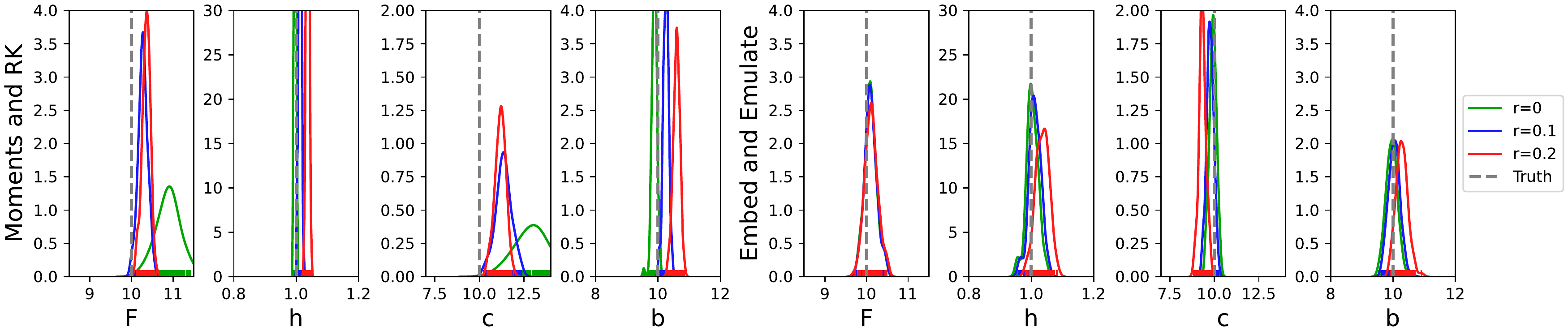}
    \caption{\textbf{Impact of observation noise.} 
    Reconstructed posterior distributions, comparing  a classical numerical solver (Runge-Kutta) plugged into EnKI to minimize \eqref{eq:moment} (left) with  \eande (right). Both variants of EnKI  are run for 70 iterations with 100 particles. The green line  shows the noise-free case, the blue shows the noisy case when $r=0.1$, and the red shows the noisy case when $r=0.2$. We see here that our proposed \eande method (right column) produces posterior estimates that are consistent over a range of noise levels, while the baseline EnKI approach using a pre-defined embedding corresponding to a moments vector is much more sensitive to variations in $r$. 
    }
    \label{fig:posterior}
\vspace{-.15in}
\end{figure}

As shown in  \cref{fig:posterior}, \eande is more robust to noise than the baseline method with numerical solvers and predefined moments. Reconstructed posterior distributions learned with \eande in \cref{fig:posterior} are more consistent with increasing noise levels, especially for the parameters $F$ and $b$.
%------------------------------------------------

\subsection{Ablation study: role of the regression head}
\label{sec:ablation}
In this section, 
we empirically verify the utility of the regression head $h_\theta$ in our \eande framework.
% \paragraph{Setup } 
We use $n=4000$ training samples in two settings: 
first, we use the full \eande model of  \cref{section:vary_train_size};
second, we  discard the regression head (i.e. fix $h_\theta = 0$) while keeping intra- and inter-domain contrastive losses. For both trained models, we run the EnKI. 
We run the EnKI using components from the \eande framework for two choices of prior: first, we use $\pfixed$, and second, we use $\pempb$. Note the empirical Bayes prior is only possible with the regression head. These priors are detailed in \cref{section:vary_train_size} and \cref{appendix:implementation}. 

Table \ref{table:ablation_regression} shows that having the regression component of the loss complement the contrastive losses yields a substantial improvement in parameter estimation accuracy using \eande. In other words, explicitly training the embedding function and emulator for the parameter estimation task  yields an emulator that is far more accurate during parameter estimation than training a generic embedding and generic emulator. Furthermore, this table illustrates the efficacy of our empirical Bayes procedure -- i.e., using our learned
 regressor to alter the prior used by EnKI. The medians are slightly improved while the means are strongly improved, suggesting that the empirical Bayes procedure is particularly helpful in the tails. 

\begin{table}[ht]
  \centering
    %   \rowcolors{2}{gray!60}{white}
  \begin{tabular}{lccccc}
    \toprule
     & {$F$  $\downarrow$} & {$h$  $\downarrow$} & {$c$  $\downarrow$} & {$b$  $\downarrow$} 
    \\
    \midrule
 (a) {\small No regression head ($\pfixed$)}  & 16.75 (2.62) & 6.02 (1.63) & 22.16 (6.35) & 6.60 (1.86) \\
 (b) {\small \eande ($\pfixed$)}  &  8.39 (1.12) & 1.77 (0.82) & 15.82 (1.61) & 3.85 (0.91) \\
 (c) {\small \eande ($\pempb$)} & \textbf{3.39 (1.03)} & \textbf{1.21 (0.76)} & \textbf{4.53 (1.52)} & \textbf{3.02 (0.90)} \\
  \bottomrule
  \end{tabular}
  \vspace{.1in}
  \caption{\textbf{Ablation study of regression head:} Average MAPE (MdAPE, median absolute percentage error)  over 200 test instances for estimating $\bphi$ by minimizing \eqref{eqn:J_obj_emu} in three different settings:
  (a) $f_\theta$ and $\hat g_\theta$  correspond to a generic emulator trained without the regression loss $\ell_{\rm reg}$ and the original prior $\pfixed$; (b)
  $f_\theta$ and $\hat g_\theta$  correspond to the emulator learned with our \eande framework and the original prior $\pfixed$; and (c) $f_\theta$ and $\hat g_\theta$  correspond to the emulator learned with our \eande framework and the empirical Bayes prior $\pempb$. 
  This experiment shows that having the regression component of the loss complement the contrastive losses yields a substantial improvement in parameter estimation accuracy using \eande. 
  }   
  \label{table:ablation_regression}
\vspace{-.1in} 
\end{table}

\section{Conclusions}
\label{sec:conclusion}
The proposed \eande framework trains an emulator of a complex simulation to facilitate parameter estimation. Unlike generic emulation methods, which can lead to poor parameter estimates and require expert domain knowledge to construct, our method (a) leverages a contrastive learning framework coupled with a regression head to jointly learn a low-dimensional embedding of simulator outputs that can be used to construct an objective function for parameter estimation and (b) yields an emulator that can be used within an optimization framework such as the EnKI to produce accurate parameter estimates in 1.19\% of the computation time of an approach using classical numerical methods, even accounting for the time required to generate training samples.  We explore our approach in the context of earth system modeling as described by \citet{schneider2017earth} and \citet{cleary2021calibrate} and hypothesize that these tools can facilitate improved climate forecasts that account for uncertainties and cover the full range of possible outcomes. The social impacts of improved climate forecasting are positive if acted upon.
While learned emulators are gaining in popularity, we as a community still know little about which systems are more or less challenging to emulate or how to design task-specific emulators more generally. There are further opportunities for exploring emulators for parameter estimation using optimization frameworks beside the EnKI. 

\section*{Acknowledgments and Disclosure of Funding}
We thank the anonymous reviewers and area chair for their helpful comments. 
We thank Owen Melia and Xiao Zhang for helpful discussion and proofreading our paper.
RJ and RW are partially supported by DOE DE-AC02-06CH113575, DE-SC0022232, AFOSR FA9550-18-1-0166, NSF OAC-1934637, NSF DMS-2023109, and NSF DGE-2022023.
% The views and conclusions contained in this paper are those of the authors and should not be interpreted as representing any funding agencies.

\clearpage

\bibliographystyle{plainnat}
\bibliography{reference.bib}
\clearpage

\appendix

\newpage
\appendix
\section{Algorithms and analysis}
\subsection{Ensemble Kalman Inversion}
\label{section:appendix_enki}
Ensemble Kalman Inversion (EnKI) gained its popularity in addressing Bayesian inverse problems since its proposal \citep{Iglesias_2013}.
It is 
a derivative-free optimization method for objective functions of the form
% way addressing
\citep{chada2020iterative}
:
\begin{equation}
    J_{\rm DM}(\bphi) := \frac{1}{2} \norm{\by_{\rm{measure}} - g(\bphi)}^2_\bR,
\end{equation}
where $\by_{\rm{measure}}$ is the measured data, $g$ is a given map and $\bR$ is a symmetric positive definite matrix representing the data measurement precision (we explain how different approaches in our paper using different $\by_{\rm measure}$ and $\bR$ below).

The EnKI estimates a posterior distribution of $\bphi$, which is approximated using an ensemble of particles. 
At each iteration, the EnKI consists of two steps: the prediction step and the analysis step. 
In the prediction step, we apply the forward model $g$ to each particle of the ensemble.
In the analysis step, each particle is updated by using artificially perturbed observations and forward-mapped particles.
The EnKI is summarized in \cref{alg:enki}.

\begin{algorithm}[h]
\caption{EnKI ($\by_{\rm{measure}}, g, p_\bphi, \alpha, \bR, M, N)$}
\label{alg:enki}
\renewcommand{\algorithmicrequire}{\textbf{Input:}}
\renewcommand{\algorithmicensure}{\textbf{Output:}}
\newcommand{\INDSTATE}[1][1]{\hspace{#1\algorithmicindent}}
\begin{algorithmic}[1]
\REQUIRE  Data $\by_{\rm{measure}}$, forward model $g$, prior distribution $p_\bphi$, step size {$\alpha$}, 
variance $\bR$,
ensemble size $M$, number of iterations $N$. \\
\INDSTATE[-2] \textbf{Initialize: } $\{\hatphi^{(0,m)}\}_{m=1}^M$ sampled from the prior distribution $p_\bphi$.
\\
\FOR {$i = 1, \cdots N$ }
 \STATE \textbf{Prediction step:} propagate the ensemble of particles to data space
 $\{g\big( \hatphi^{(i, m)} \big)\}_{m=1}^M$ by applying the forward model $g$ $M$ times. 
 From the ensemble, calculate the empirical means and covariance matrices:
 	$$
    \xoverline{\bphi}^i = \frac{1}{M}\sum_{m=1}^M \bphi^{(i, m)}, \text{ }
 	\xoverline{g}^i = \frac{1}{M}\sum_{m=1}^M g\big(\hatphi^{(i, m)}\big),	
 	$$
 	$$ 
 	\bC^{i}_{\bphi g} = \frac{1}{M}\sum_{m=1}^M \Big( \hatphi^{(i, m)}_i -\xoverline{\bphi}^i \Big) \otimes \Big( g\big(\hatphi^{(i, m)}\big) - \xoverline{g}^i \Big),	 
 	$$
 	$$ 
 	\bC^{i}_{g g} = \frac{1}{M}\sum_{m=1}^M \Big( g\big(\hatphi^{(i, m)}\big) - \xoverline{g}^i \Big) \otimes \Big( g\big(\hatphi^{(i, m)}\big) - \xoverline{g}^i \Big),	 
 	$$
 \STATE \textbf{Analysis step}: calculate the Kalman gain matrix 
 $$\bK^i = \bC^{i}_{\bphi g} (	\bC^{i}_{g g} + \alpha^{-1} \bR)^{-1},$$
 artificially perturb the data $\by_{\rm{measure}}^{(i,m)} = \by_{\rm{measure}} + \eta^{(i,m)}$ where $\eta^{(i,m)} \sim \mathcal{N}(0, \alpha^{-1}\bR)$,
 and update the ensemble of particles:
 $$
 \hatphi^{(i+1, m)} =  \hatphi^{(i, m)} + \bK^i \Big(\by_{\rm{measure}}^{(i,m)} - g\big(\hatphi^{(i, m)}\big) \Big), 1 \leq m \leq M.
 $$
\ENDFOR
\ENSURE $\{ \hatphi^{(N, m)} \}$
\end{algorithmic}
\label{appendix:enki_algorithm}
\end{algorithm}

When using \eande to minimize the objective function \eqref{eqn:J_obj_emu}, $\by_{\rm{measure}}$
corresponds to the embedding of the raw trajectory $f_\theta(\bZ)$, the forward model $g$ corresponds to the emulator $\hat{g}_\theta$, and $\bR$ corresponds to an identity matrix. 
When using the baseline method to minimize the objective function \eqref{eq:moment}, $\by_{\rm{measure}}$ 
corresponds to the moment vector $m(\bZ)$, $g$ corresponds to the $m \circ H$ where $H$ is computed using Runge-Kutta method, and $\bR$ is a diagonal matrix with $j$-th diagonal entry $\bR_{j, j}:= {\rm Var}[ m(\bZ)_j]$.

% _____________________________________
\subsection{Contrastive learning and inverse problems}

\label{appendix:theoretical_analysis}
In this section, we briefly reviewed the assumptions and results of
\citet{zimmermann2021contrastive} and explain how we relate the analysis to our work in addressing the parameter estimation problem.

Recent literature studies the reason why contrastive learned representations can be successfully extended to multiple downstream tasks \citep{arora2019theoretical, wang2020understanding, zimmermann2021contrastive}.
Among them, a recent work from \citet{zimmermann2021contrastive} points out that under certain assumptions, the representations $f_\theta$ learned in the contrastive framework to minimize $\ell_{\bZ\bZ}(\theta; \tau)$ (shown in \eqref{eqn:lcontra_f}) inverts the underlying data generating process $H^{-1}(\bZ)$ of observed data up to an affine transformation.

\paragraph{Assumptions} \citet{zimmermann2021contrastive} firstly make assumptions on the  \textit{data generating process}. They assume the generator $H: \Phi \rightarrow \mathcal{Z}$ is an injective function with $ \Phi \in \mathbb{S}^{k-1}$  in a unit hypersphere and the distribution of $\bphi \in \Phi$ is uniform.
Second, they assume the conditional distribution $p(\tilphi|\bphi)$ of a ``positive parameter'' $\tilphi$ given $\bphi$ is a von Mises-Fisher (vMF) distribution
with $p(\tilphi|\bphi) = c_p^{-1} \exp(\kappa \ang{\tilphi, \bphi})$ where $c_p$ is the  normalizing constant
% $c_p:= \int \exp( \kappa \ang{\bphi, \tilphi}) {\rm d} \tilphi$ 
and $\kappa$ is the concentration parameter.
Third, they assume the learned representation $f_\theta: \mathcal{Z} \rightarrow \mathbb{S}^{k-1}$ is in the unit hypersphere where the dimension of the representation is equal to the dimension of the parameters.
They further assume the learned conditional distribution of $\tilphi$ given $\bphi$ with the contrastive trained neural network is a vMF distribution with $q_\theta(\tilphi|\bphi) = C_q(\bphi)^{-1} \exp ( \ang{f_\theta(H(\tilphi), f_\theta(H(\bphi)) / \tau})$ where $C_q(\bphi)$ is the partition function.

In the context of the parameter estimation problem of this paper, the relationship proved by \citet{zimmermann2021contrastive}, that $f_\theta(\bZ) = A H^{-1}(\bZ) = A\bphi$ where $A$ is a full rank matrix, is clearly a desirable  property as,
if $A$ were known, 
we could simply invert $A$ 
to obtain an unbiased estimate of $\bphi$.
Empirically, we find that directly adding a linear regression head on top of $f_\theta$ (which effectively serves as a mechanism for learning $A^{-1}$; see \cref{fig:diagram_framework}) provides faster convergence and better performance (as shown in \cref{sec:ablation}).

% _____________________________________

\section{Implementation details}
\subsection{Experiment details}
\label{appendix:implementation}

\paragraph{Lorenz 96 dataset}
Initial conditions of the ODE/PDE systems affect the values of dynamical variables at future times.
To better fit in real-world scenarios, we simulate the training data used for \eande and the supervised regression, as well as the testing data used for evaluations of all methods, with random initial conditions sampled from the normal distributions. 
During EnKI evaluations of the baseline
we are required to specify initial conditions.
We follow \citet{schneider2017earth} and set the initial conditions of all simulations in the EnKI evaluations as random samples drawn from the 
observation.
Note that, unlike the baseline method, \eande directly learns the structural embeddings of $\bphi$ and does not require initial conditions.
Beyond the setup of initial conditions, there are two stages to generate our train and test data. In the first stage, we generate a collection of $\bphi_i$, and for each one, we run a Runge-Kutta solver to compute a corresponding $\bZ_i$.
In the second stage, we filter out some of the $\{\bphi_i, \bZ_i\}$ generated during the first stage --whenever the Runge-Kutta iterations are unsuccessful, generating NaN values, and whenever the $\bZ_i$ are degenerate, resulting in a standard deviation below 
5e-5.

\paragraph{Data processing } 
(a)\textit{Trajectory cropping:} during the training of \eande and the supervised regression, we randomly crop
a collection of length-250 trajectories from length-1000 trajectories, where each crop $\bZ_{\rm{crop}} \in \mathbb{R}^{250\times 396}$.
(b) \textit{``positive'' data selection:}
\eande trained with contrastive learning objectives tries to pull ``anchor'' data $\bZ$ (or $\bphi$) towards ``positive'' data $\btilZ$ in \eqref{eqn:lcontra_f} (or $\tilphi$ in \eqref{eqn:lcontra_g}) and at the same time push ``negative'' data away.
To identify a ``positive'' trajectory $\btilZ$ (or parameter $\tilphi$),
we use a two-step approach.
First, we select it in a supervised manner by measuring the distance between the ``anchor'' data $\bZ_i$ (or $\bphi_i$) and a sample $\bZ_j$ (or $\bphi_j$) in the parameter space.
We find the index $\tilde{i}$ of the nearest neighbor of the ``anchor'' data in the parameter space with $\tilde{i} := \arg \min_{i \neq j} \delta (\bphi_i, \bphi_j)$, where $\delta$ is a metric function defined as: $\delta(\bphi_i, \bphi_j) := \frac{1}{2} \big\{ {\rm APE}(\bphi_i; \bphi_j) + {\rm APE}(\bphi_j; \bphi_i)\big\} $ (see $\rm{APE}$ in \cref{sec:appendix_perform_metric}). 
However, since the training data is sampled from finite grids with limited size, the distance between the ``anchor'' data and its nearest neighbor $\delta(\bphi_i, \bphi_{\tilde{i}})$ could be very large.  
We apply a threshold filter and only accept a candidate nearest neighbor as ``positive'' data when the empirical distance $\delta (\bphi_i, \bphi_{\tilde{i}})$ is below some threshold (e.g., $0.45$ for $n=500$, $0.4$ for $n=1,000$ and $4,000$ ). 
Second, for the cases where no sample passes the threshold filter, we set $\btilZ$ and $\tilphi$ in two different ways.
For $\btilZ$, we simply set it in an unsupervised way by randomly making another crop on $\bZ_i$.
For $\tilphi$, we set it by randomly adding a small amount of 
% multiplicative 
noise on $\bphi_i$ with $\tilphi_{ij} := \bphi_{ij} + \xi \bphi_{ij}, j=1,2,3,4$ with a certain probability (e.g., $50\%$), 
where $\xi$ is a normal variable with mean 0 and a small standard deviation (e.g., 0.04).

\paragraph{Training Hyperparameters} We train \eande and the supervised regression using the AdamW optimizer. For both methods, we linearly warm up the learning rate to 0.01 at the beginning of the training and then gradually decay the learning rate with a cosine decay scheduler. 
We use $\lambda = $1e-5 for weight decay.
We train NPE-C \citep{greenberg2019automatic} and SNL+ \citep{chen2020neural} with learning rate 1e-4 and 5e-4 as we find lower learning rate leads to better performance.
We train all the methods with $1,000$ epochs.
We use batch size $1000$ for large sized training data with $n \geq 1,000$. 
For the small size training data with $n = 500$, we set it as $500$.

Temperature values in contrastive learning balance the influence between barely distinguishable samples and easily distinguishable ``negative'' samples \citep{ravula2021inverse}. They are commonly chosen to be less than one where smaller values indicate a larger influence of barely distinguishable     ``hard'' samples. For \eande, we initialize the temperature values $\tau$ in \eqref{eqn:lcontra_f} and \eqref{eqn:lcontra_g} and $\tau'$ in \eqref{eqn:lcontra_gf} at $0.15$. We keep $\tau$ and $\tau'$ fixed at $0.15$ for the first $500$ epochs to make sure ``hard'' samples get large gradients and are updated sufficiently.
We then adopt the ``heating-up'' strategy from \citet{zhang2018heated} and linearly increase $\tau$ in \eqref{eqn:lcontra_f} and \eqref{eqn:lcontra_g} to 0.5 to increase the impact of ``easier'' samples. We keep $\tau'$ in \eqref{eqn:lcontra_gf} fixed all the time so that the learned emulator $\hat{g}_\theta$ is capable of distinguishing embeddings of ``hard'' samples from the embedding network $f_\theta$, and vice versa. 

In training a contrastive learning target, we usually need sufficient amount of ``negative'' samples to approximate the uniform repulsion. We follow \citet{he2020momentum} and construct a first-in-first-out queue with a rolling updating scheme. Practically, the memory bank size should be no less than the training size. In our experiments, we set the size of the memory bank as $5,000$ when the training size $n=4,000$; 
% the size of the memory bank as $3,000$ when the training size $n=2,000$; 
the size of the memory bank as $2,000$ when the training size $n=1,000$; 
and the size of the memory bank as $1,500$ when the training size $n=500$.

All the hyperparameters are chosen using a grid search with a reserved validation set consisting of 5 samples.  The range of values searched over are as follows:
\begin{squishlist}
    \item The initialized learning rates for \eande and the supervised regression were selected from the set $\{0.001, 0.01\}$.
    \item The initialized temperature values of $\tau$ in \eqref{eqn:lcontra_f} and \eqref{eqn:lcontra_g} and $\tau'$ in \eqref{eqn:lcontra_gf} were selected from the set $\{0.10, 0.15, 0.20\}$. The heated up maximum value of $\tau$ in \eqref{eqn:lcontra_f} and \eqref{eqn:lcontra_g} were selected from the set $\{0.1, 0.2, 0.3, 0.4, 0.5\}$.
\end{squishlist}

\paragraph{Priors Initialization of the EnKI} 
As \citet{schneider2017earth}, for the initialization of the EnKI prior,
we use a normal prior for $(\bphi_1, \bphi_2, \bphi_4) = (F, h, b)$ and a log-normal prior for $\bphi_3 = c$ in order to enforce positivity.
For the baseline method minimizing \eqref{eq:moment} in \cref{section:vary_train_size} and \cref{sec:exp_noisy}, we use a fixed prior $p_{\bphi, {\rm fixed}}$.
For the normal variables, we choose the mean at the middle of the range of testing instances: 
$\mu_{\bphi, {\rm fixed}}(F, h, b) = (7.5, 2.5, 12.5)$ 
% $\mu_{\bphi, {\rm fixed}}(F, h, b) = (8.0, 2.5, 11.5)$ 
and variances broad enough so that all test samples are within $2\sigma$ of the mean: $\sigma^2_{\bphi, {\rm fixed}}(F, h, b) = (36.0, 2.25, 36.0)$. 
For the log-normal variable $\log c$, we set its mean $\mu_{\bphi, {\rm fixed}}(\log c) = \log(11.5)$ and variance $\sigma^2_{\bphi, {\rm fixed}}(\log c) = 0.15$ (i.e., a mean value of $12.5$ for $c$).
For \eande, we can choose between the instance-wise prior $p_{\bphi, {\rm empB}}$ and the fixed prior defined above. Specifically, we alter the mean of $p_{\bphi, {\rm empB}}$ with the empirical estimate provided by the regression head and use a relatively small variance. 
$\sigma^2_{\bphi, {\rm emp}}(F, h, b) = (18.0, 1.125, 18.0)$ for normal variables and $\sigma^2_{\bphi, {\rm emp}}(\log c) = 0.075$ for log-normal variable $c$.

\paragraph{Setup of the EnKI} 
For experiments in \cref{section:vary_train_size}, we set the ensemble size $M=100$, step size $\alpha = 0.3$ and iteration number $N=50$ for the baseline method with $p_{\bphi, {\rm fixed}}$ and \eande with $p_{\bphi, {\rm empB}}$. 
For the study of noise impact in \cref{sec:exp_noisy} when evaluating the baseline and \eande, we use $p_{\bphi, {\rm fixed}}$.
% , we set the step size $\alpha = 0.5$.
For the ablation study in \cref{sec:ablation} when evaluating \eande and no regression head contrastive learning with $p_{\bphi, {\rm fixed}}$, we set a large ensemble size $M = 10,000$ to prevent potential collapse in a local minimum and $N = 100$ to ensure convergence. Arguably, this could slow down the computational times. However, empirically we only increased the time from $2$ seconds to $16$ seconds per instance, which is significantly lower than the times consumed by the baseline method.

\paragraph{Computational resources} Experiments of \eande and the supervised regression were performed on a system with 4x Nvidia A40 GPUs, 2 AMD EPYC 7302 CPUs, and 128GB of RAM. The baseline methods and data generating process were performed on 2x Xeon Gold 6130 CPUs.

\subsection{Network Architecture}
\label{appendix:nn_architecture}
\begin{figure}[ht]
    \centering
    \subfloat[Encoder network]{\includegraphics[width=.2\linewidth]{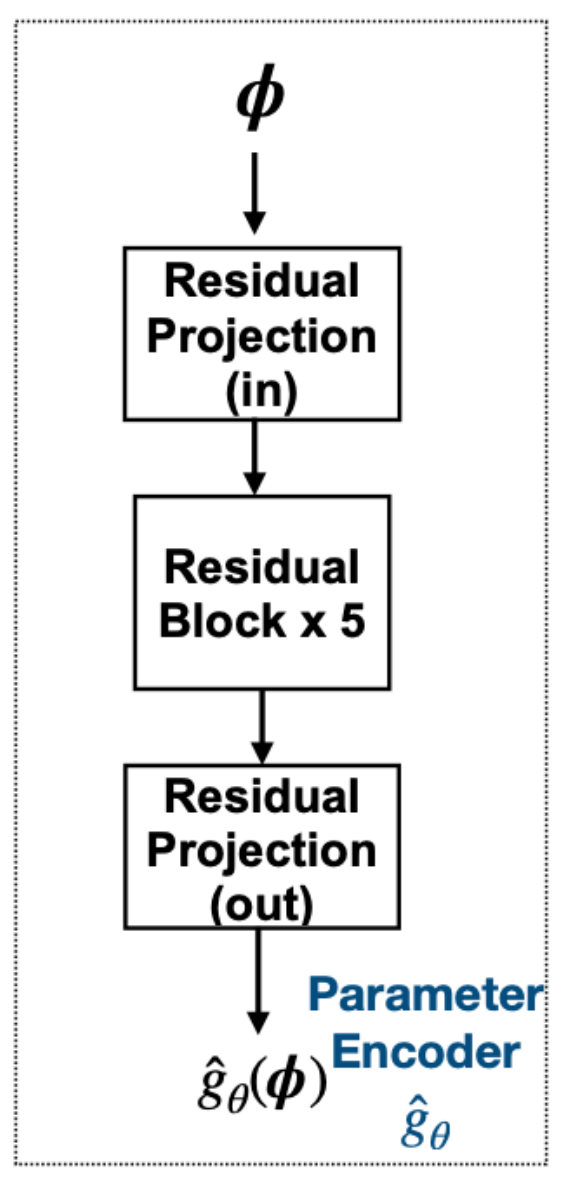}}
    \subfloat[Residual projection (in)]{
    \includegraphics[width=.25\linewidth]{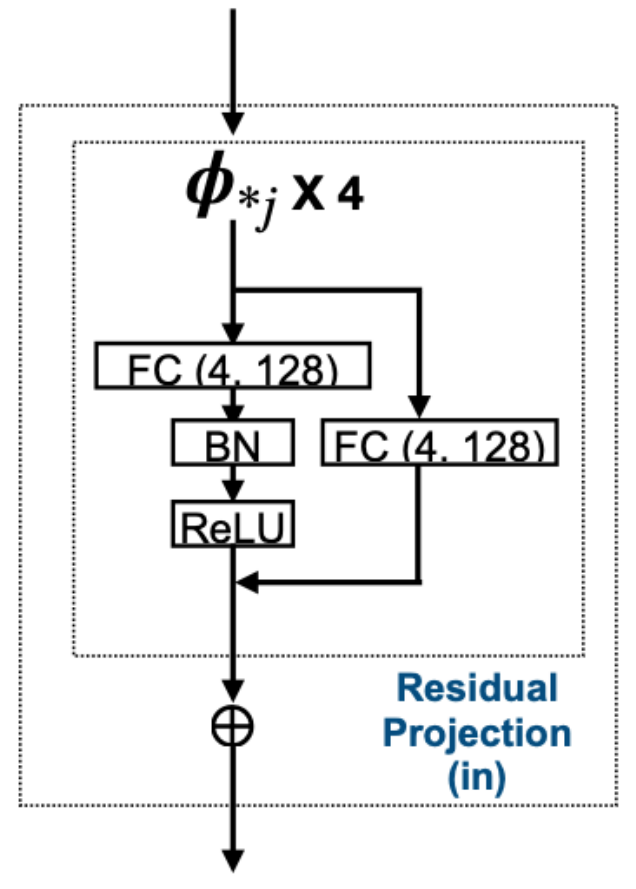}}
    \subfloat[Residual block]{
    \includegraphics[width=.2\linewidth]{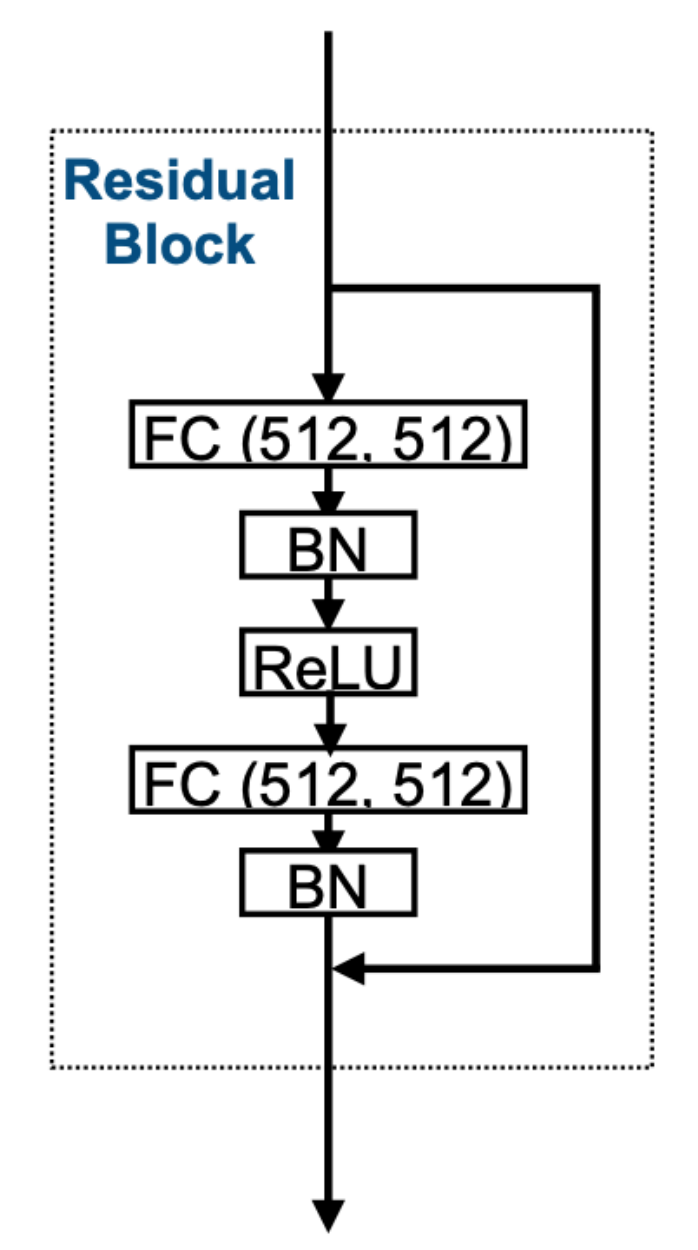}}
    \subfloat[Residual projection (out)]{
    \includegraphics[width=.25\linewidth]{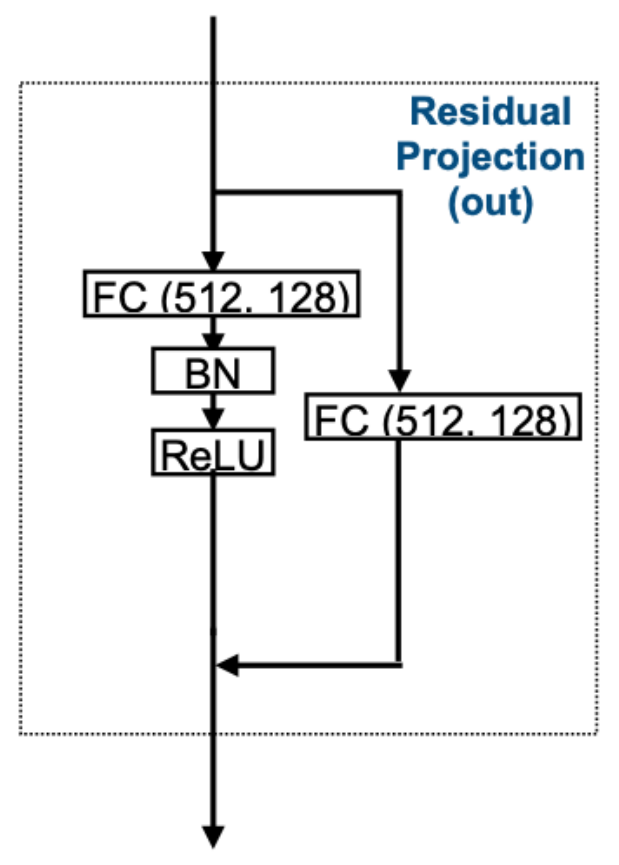}}
    \caption{\textbf{Architecture of emulator $\hat{g}_\theta$ of \eande} (a) The emulator network consists of one residual projection (in), three residual projection blocks, and one residual projection (out). 
    (b) The residual projection (in) independently maps the $j$-th item of $\bphi$ into a high-dimensional (e.g., 128) latent representation, which is then concatenated (shown as the $\oplus$ operator) as a single vector as the output.
    (c) The residual block has the same input and output dimension. 
    (4) The residual projection (out) projects a high-dimensional input latent vector into a low-dimensional embedding.}
    \label{fig:diagram_emulator}
\end{figure}

\paragraph{Emulator $\hat{g}_\theta$}To enhance the representation power of the emulator $\hat{g}_\theta$ and enlarge the alignment between two learned representation spaces, we design the emulator using residual blocks as in \cref{fig:diagram_emulator}.

\paragraph{Embedding network $f_\theta$ } We use Resnet34 \citep{he2016deep} as the backbone of the trajectory branch of \eande in \cref{fig:diagram_framework}(a) and the supervised regression. We apply average pooling at the last layer to generate a 512-d hidden vector.
To respect the temporal periodic pattern of Lorenz 96, we replace the standard zero padding with the circular padding on the temporal axis.
For the regression head $h_\theta$, we project the 512-d hidden vector into a 4-d output using only a single fully connected layer. 
For the embedding network, we project the 512-d hidden vector into a 128-d feature vector using a similar structured branch as projection residual (out) in \cref{fig:diagram_emulator}(d) except that the residual block skips two layers instead of one.

\subsection{Performance metrics}
\label{sec:appendix_perform_metric}

\paragraph{Absolute Percentage Error (APE)}:  APE is used in selecting ``positive'' samples of \eande and is calculated 
on pairs of training samples:
\begin{equation}
    {\rm APE}(\bphi_j; \bphi_i) = \sum_{l=1}^{k} \frac{|\bphi_{il} - \bphi_{jl}|}{|\bphi_{il} + \epsilon|},
\end{equation}
here $k$ is the dimension of $\bphi$, $i, j$ represent indexes of two data samples, and $\epsilon$ is a small value to address overflow issues.

\paragraph{Mean Absolute Percentage Error (MAPE)} MAPE is a measure of prediction accuracy over the entire testing data and is expressed as a ratio with the formula:
\begin{equation}
 {\rm MAPE}(\hatphi_{*j}; \bphi_{*j}) = \frac{1}{n}\sum_{i=1}^{n}\frac{|\hatphi_{ij} - \bphi_{ij}|}{|\bphi_{ij}| + \epsilon},
\end{equation}
where $\hatphi_{*j}$ represents the $j$-th component of the estimate parameter vector,
$\bphi_{*j}$ represents the $j$-th component of the true parameter vector (e.g., $[\bphi_{*1},\bphi_{*2}, \bphi_{*3}, \bphi_{*4} ] = [F, h, c, b]$ for Lorenz 96),
and $\epsilon$ is a small value to address overflow issues.

\paragraph{Continuous Ranked Probability Score(CRPS)} 
As described in \cref{section:vary_train_size}, CRPS is a probability metric to evaluate the performance of a probabilistic estimation. CRPS is computed as a quadratic measurement of cumulative distribution function (CDF) between the predicted distribution $C$ and the empirical distribution of the observation. CRPS is defined as \citep{hersbach2000decomposition, zamo2018estimation, pappenberger2015know}:
\begin{equation}
    {\rm CRPS}(C, \bphi_{*j}) = \int^{\infty}_{-\infty}(C(y_j) - U(y_j - \bphi_{*j}))^2 {\rm d}y_j,
    \label{eqn:crfs_integrate}
\end{equation}
where $\bphi_{*j}$ and $\hatphi_{*j}$ are defined above,
$C$ is the CDF of the ensemble estimation with $C(y_j) = P(\hatphi_{*j} \leq y_j)$,
and $U$ is the CDF of the Heaviside step function with $U(y_j - \bphi_{*j}) = \mathbf{1}\{y_j \geq \bphi_{*j} \}$. Practically estimating $C$ is difficult since it has no analytic form. 
In our case, using EnKI with ensemble size $M$, we can estimate \eqref{eqn:crfs_integrate} 
using the empirical distribution calculated from the particles $\{\hatphi_{*j}^{(m)}\}_{m=1}^M$\footnote{\url{https://pypi.org/project/properscoring/}}:
\begin{equation}
    {\rm CRPS} (C, \bphi_{*j}) = -\frac{1}{2M^2} \sum_{m=1}^M\sum_{m' =1}^M|\hatphi_{*j}^{(m)} - \hatphi_{*j}^{(m')}| 
    +
    \frac{1}{M}\sum_{m=1}^M|\hatphi_{*j}^{(m)} - \bphi_{*j}|.
\end{equation}

\subsection{Visualizing the objective functions}
\begin{figure}[ht]
    \centering
    \subfloat[Marginal heatmap of \eqref{eq:moment} using Runge-Kutta]{\includegraphics[width = .5\linewidth]{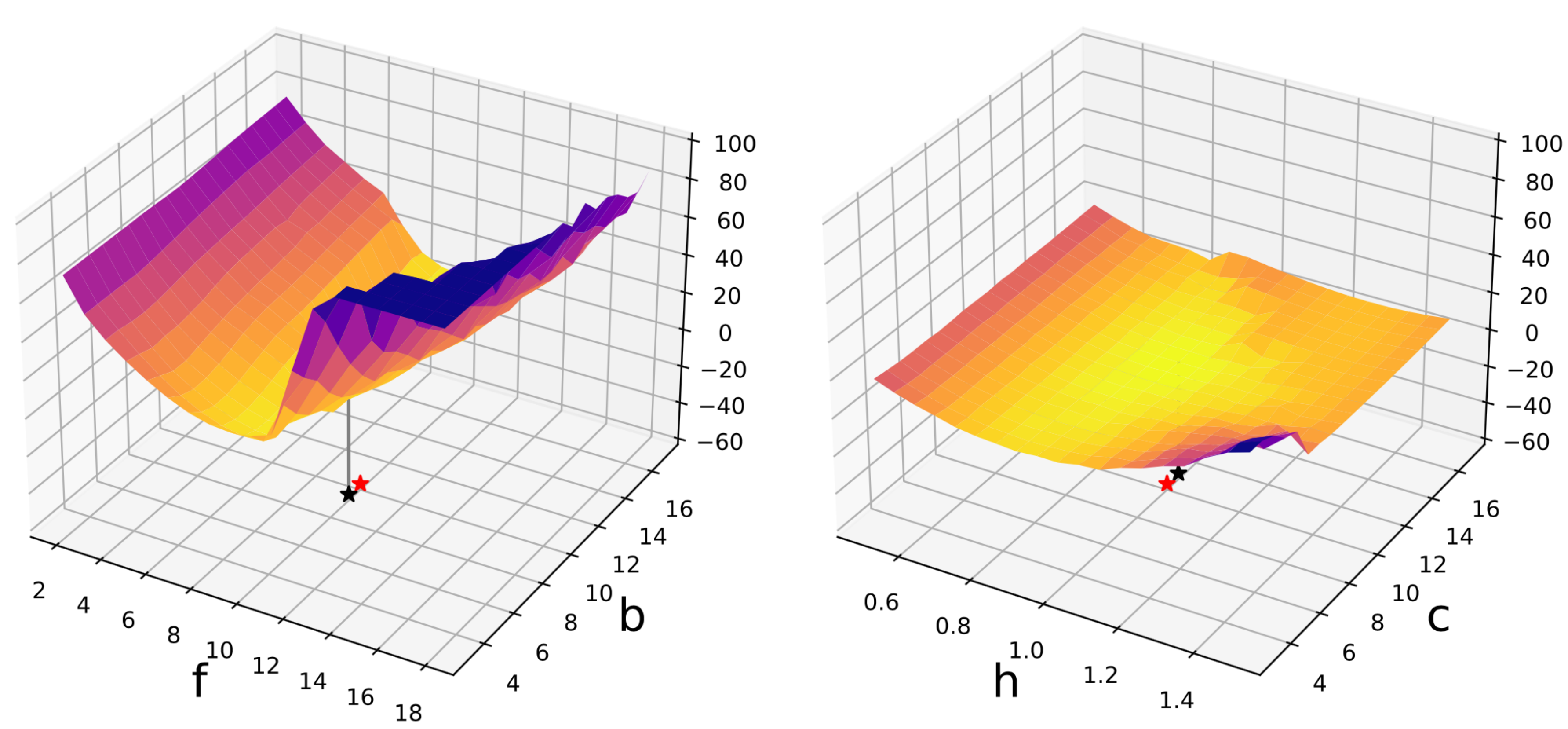}}\hfill
    \subfloat[Marginal heatmap of \eqref{eqn:J_obj_emu} using learned emulator.]{\includegraphics[width =.5\linewidth]{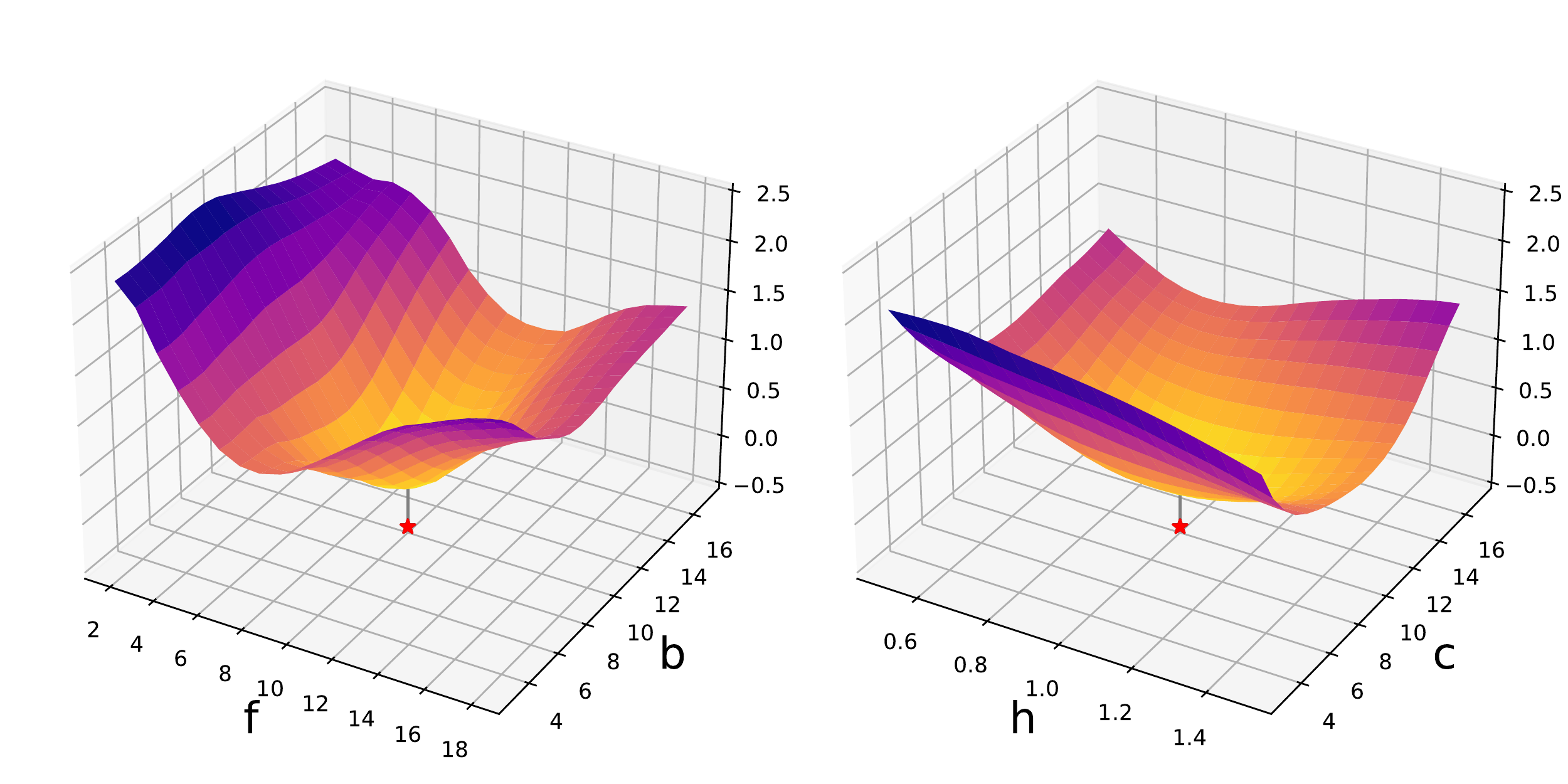}}
\caption{\textbf{Heatmap visualization showing values of objective functions.} (a) the marginal heatmap of predefined moments objective function \eqref{eq:moment}; and (b) the marginal heatmap of \eqref{eqn:J_obj_emu} with \eande using learned emulator. 
The red stars in both plots show the locations of the true parameters, while the black stars show the locations of the points with the minimum function value.
All objective values are greater or equal to zero, but we intentionally set the limit of $z$-axis to be negative for clear visualization of ``star'' points.
$z$-axes between different marginal heatmaps using the same objective function are aligned with post-processing.
% Values of the objective function \eqref{eqn:J_obj_emu} are bounded between $[0,2]$ since the embeddings are in a unit hypersphere.
Unlike the value of the objective function \eqref{eqn:J_obj_emu} are bounded between $[0,4]$, 
values of \eqref{eq:moment} have no upper bound and might contain extremely large numbers. 
For better visualization, we manually truncate values of \eqref{eq:moment} by a threshold number (e.g.,100) for better visualization.}
\label{fig:vis_objective}
\end{figure}
% \roxie{new edits in this section}
In this section, we visualize the objective function of \eqref{eq:moment} with predefined moments function and Runge-Kutta and \eqref{eqn:J_obj_emu} of \eande with emulator $f_\theta$ and $\hat{g}_\theta$. 
We use the same instance in \cref{sec:exp_noisy} as the observation with the ground truth $\bphi = [10, 1, 10, 10]$ and the noiseless $\bZ$ is of length $10,000$.
\cref{fig:vis_objective} shows the marginal heatmaps of the objective functions when holding two parameters fixed at the ground truth values and sampling the other two parameters uniformly in a fixed range.
We can see that the objective function learned with \eande can accurately reflect the dynamics in the parameter space with smooth curves.
In contrast, the predefined objective function is insensitive to the parameter $c$ with a flat curve showing in the marginal heatmap of $(h,c)$ pairs.

\section{Further Experiments}
In this section, we study the 1-d Kuramoto-Sivashinsky equation (KSE).
Kuramoto-Sivashinsky equation is known for its chaotic dynamics and can be applied in describing a variety of physical phenomena, including flows in pipes, plasmas, chemical reaction dynamics, etc. \citep{hyman1986kuramoto}
The Kuramoto-Sivashinsky equation is a fourth-order partial differential equation with the form:
\begin{eqnarray*}
    \pdv{\bV^x_t}{t} + \lambda_2\pdv[order=2]{\bV^x_t}{x} + 
    \lambda_4\pdv[order=4]{\bV^x_t}{x} +
    \lambda_{\rm nonlinear} \bV^x_t \pdv{\bV^x_t}{x} = 0, \\
    \text{and } \bV^x_t = \bV^{x+2L}_t,
\end{eqnarray*}
where $\bV^x_t$ is the state of KSE, $\lambda_2$, $\lambda_4$ and $\lambda_{\rm nonlinear}$ are the coefficients of second-order derivative, fourth-order derivative and nonlinear term.
In all, we define $\bphi := [\lambda_2, \lambda_4, \lambda_{\rm nonlinear}]$, and our goal is to learn $\bphi$ from sequences of observations $\bV \in \mathbb{R}^{T \times d}$.

\paragraph{Choosing the moments function}
We choose the first order statistics of $\bV$ to define the moments function: $m(\bV) = [\ang{\bV}]$ where $\ang{\cdot}$ denotes empirical average over time. 
Note here we have tried multiple combinations of moments, including second-order and third-order statistics but found no choice of moments performs better than using only the first-order. 

\paragraph{Preparation of dataset} 
We set $L = 32$ and $d = 256$ to ensure we are in a chaotic regime.
Initial conditions are randomly sampled from a uniform range $[-\pi, \pi]$.
All simulations performed here are completed with a Runge-Kutta 4-stage method as \citet{pachev2021concurrent}.
Training data are sampled uniformly for all three components of $\bphi$ ranging from $\min = 0.1$ to $\max = 10.0$ with length $T=400, dt = 0.5$ and size $n=500$.
100 testing data are sampled in a narrower range from $\min = 0.5$ and $\max = 9.5$ with length 800 and $dt=0.5$. 

\paragraph{Setup of the EnKI} We set the fixed Gaussian prior $p_{\bphi, {\rm fixed}}$ with mean at the middle of the testing range and variance broader enough to cover all testing instances in $2 \sigma$: $\mu_{\rm \bphi_{*j}, fixed} = 5, \sigma^2_{\rm \bphi_{*j}, fixed} = 6.25, j=1, 2, 3$.
We use the same variance for $p_{\bphi, {\rm empB}}$ and alter its mean with the empirical estimate provided by the regression head.
For both baseline and \eande, we set ensemble size $M = 100$, iteration number $N = 40$, step size $\alpha = 0.3$.

\paragraph{Setup of the training} We train both \eande and the supervised regression for 1,000 epochs. We set the memory bank size of \eande for 1,000.
We set the dimension of learned representation, i.e., $f_\theta(\bV)$ and $\hat{g}_\theta(\bphi)$ to be 120.
We use the same learning rate and cosine decay scheduler. 
We initialize $\tau, \tau' = 0.1$, and linearly warm up $\tau'$ from 500 to 900 epochs with the maximum value $0.6$.
 
\begin{table}[ht]
  \centering
    %   \rowcolors{2}{gray!60}{white}
  \begin{tabular}{lcccc}
    \toprule
     & {$\lambda_2\downarrow$} & {$\lambda_4 \downarrow$} & {$\lambda_{\rm nonlinear}\downarrow$} 
    \\
    \midrule
 (a)  EnKI \textit{w/o} Learning &34.73(25.44) &  56.14 (38.27) & 68.64(33.90)&  \\
 (b) $h_{\theta}$ \textit{w/} \eande  & \textbf{3.24} (1.56) &  3.70 (1.68) & \textbf{2.69} (1.45) \\
 (c) EnKI \textit{w/} \eande  & 3.48 (1.42) &  \textbf{3.29} (1.50) & 3.30 (1.69) \\
 (d) Supervised regression  & 3.27  (\textbf{1.16}) & 3.34 (\textbf{0.97}) & 3.14 (\textbf{1.36})  \\
%  (e) NPE-C  & 21.44  (13.26) & 23.76  (13.59) & 21.23 (12.50)  \\
 (e) NPE-C  & 4.30 (2.14) & 4.86 (2.59) & 3.30 (1.83) \\
  \bottomrule
  \end{tabular}
  \vspace{.1in}
  \caption{\textbf{Averaged MAPE (MdAPE, median absolute percentage error) over 100 test instances}. (a) EnKI plugged in the Runge-Kutta solver and predefined moments function in minimizing \eqref{eq:moment}; (b) \eande with Regresssion head ($h_\theta$); and 
  (c) EnKI plugged in our \eande with the empirical Bayes prior $\pempb$;
  (d) Supervised regression; 
  (e) NPE-C \citep{greenberg2019automatic}.
  } 
  \label{table:mape_kse}
\end{table}

\begin{table}[ht]
  \centering
  \begin{tabular}{lcccc}
    \toprule
     & {$\lambda_2\downarrow$} & {$\lambda_4 \downarrow$} & {$\lambda_{\rm nonlinear}\downarrow$} 
    \\
    \midrule
 (a) EnKI \textit{w/o} Learning & 1.14 & 1.50 & 1.47 \\
 (b) $h_{\theta}$ \textit{w/} \eande  & 0.148 &  0.149 & 0.125 \\
 (c) EnKI \textit{w/} \eande  & \textbf{0.121} &  \textbf{0.101} & \textbf{0.114} \\
 (d) Supervised regression  & 0.149 & 0.124 &   0.150\\
 (e) NPE-C  & 0.175 & 0.217 & 0.131   \\
  \bottomrule
  \end{tabular}
  \vspace{.1in}
  \caption{\textbf{Averaged continuous ranked probability score (CRPS) over 100 test instances}. (a) EnKI plugged in the Runge-Kutta solver and predefined moments function in minimizing \eqref{eq:moment}; (b) \eande with Regresssion head ($h_\theta$); and 
  (c) EnKI plugged in our \eande with the empirical Bayes prior $\pempb$;
  (d) Supervised regression; 
  (e) NPE-C \citep{greenberg2019automatic}.
  }   
  \label{table:crps_kse}
\end{table}

\begin{table}[ht]
  \centering
  \begin{tabular}{p{12em}||p{5em}p{4em}p{4em}p{7em}}
   %{clccccc}
    \toprule
     & {\small Training data generation} & {\small Training Time} & {\small 100 testing runs} & {\small Total time(training+ testing)} 
    \\
    \midrule
 {\small EnKI \textit{w/o} Learning}  & 0.0 & 0.0  & 3,600.0  & 3,600.0 (60h) \\
{\small $h_\theta$  \textit{w/} \eande}   &  9.0 & 34.0 & 0.5 & 43.5 (0.73  h) \\
{\small EnKI  \textit{w/} \eande}   &  9.0 & 34.0 & 1.0 & 44.0 (0.74  h) \\
{\small Supervised Regression}   & 9.0 &  28.0 & 0.5 & 39.5 (0.62 h) \\
{\small NPE-C} & 9.0 & 31.0 & 1.0 & 40.0 (0.66 h) \\
\bottomrule
\end{tabular}
\vspace{.1in}
  \caption{\textbf{Empirical computational time for different stages of different methods}. Reported in minutes, total time needed for \eande or the supervised regression are $1.23\%$ of EnKI with Runge-Kutta.
  }   
  \label{table:computational_time_kse}
\end{table}

\paragraph{Results}  We first evaluate the accuracy of point estimates using the mean absolute percentage error (MAPE) and the median absolute percentage error (MdAPE).
In Table \ref{table:mape_kse}, we see that \eande significantly outperforms EnKI using the predefined moments function and the expensive numerical solver.
This experiment demonstrates that designing moments function like \citet{schneider2017earth} did for the Lorenz 96 system needs physical information of the dynamical system
and in applications (e.g., climatology) this requires extra domain expertise.
We then evaluate the quantified uncertainty using the continuous ranked probability score (CRPS). 
In Table \ref{table:crps_kse}, we see that \eande outperforms the baselines of the EnKI without learning and the supervised regression. 
This experiment demonstrates that \eande with the EnKI is capable of providing a reliable probabilistic estimation with low errors than the EnKI without learning which requires domain knowledge to perform well, the supervised regression which only provides points estimates, 
and NPE-C which lacks competence when trying to estimate multiple different observations at test time.
Lastly, we show the empirical computational time needed for \eande, the supervised regression, NPE-C, and the EnKI with Runge-Kutta for different stages.
From Table \ref{table:computational_time_kse}, we see that \eande, the supervised regression, and NPE-C only require $1.23\%$ (or below) of the time needed by the EnKI with Runge-Kutta.

% Optionally include extra information (complete proofs, additional experiments and plots) in the appendix.
% This section will often be part of the supplemental material.

\end{document}